\pdfoutput=1
\documentclass[10pt,twocolumn,letterpaper]{article}

\usepackage[pagenumbers]{cvpr}    

\usepackage{graphicx}
\usepackage{scalerel}
\usepackage{amsmath}
\usepackage{amssymb}
\usepackage{booktabs}
\usepackage{caption}
\usepackage{subcaption}
\usepackage{float}
\usepackage{placeins}
\usepackage{stfloats}
\usepackage{enumitem}
\usepackage{tabularx}
\usepackage{xspace}
\usepackage{url}
\usepackage{xcolor}
\usepackage[hang,flushmargin]{footmisc}
\usepackage{multirow}
\usepackage{arydshln}
\newcommand\blfootnote[1]{%
  \begingroup
  \renewcommand\thefootnote{}\footnotetext{#1}%
  \endgroup
}

\newcommand\rurl[1]{%
  \href{https://#1}{\nolinkurl{#1}}%
}

\newcommand{\moonstep}{\protect\scalerel*{\includegraphics{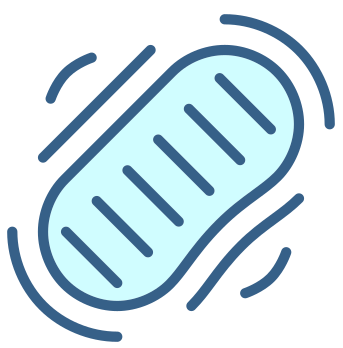}}{H}}

\definecolor{cvprblue}{rgb}{0.21,0.49,0.74}
\usepackage[pagebackref,breaklinks,colorlinks,allcolors=cvprblue]{hyperref}

\usepackage[capitalize]{cleveref}
\crefname{section}{Sec.}{Secs.}
\Crefname{section}{Section}{Sections}
\crefname{table}{Table}{Tables}
\crefname{figure}{Fig.}{Figs.}
\crefname{appendix}{App.}{Apps.}

\def\paperID{17570}
\def\confName{CVPR}
\def\confYear{2026}

\title{\moonstep{} One Small Step in Latent, One Giant Leap for Pixels:\\
Fast Latent Upscale Adapter for Your Diffusion Models}

\vspace{-3mm}
\author{
\noindent
\begin{minipage}[t]{0.3\textwidth}
    \centering
    Aleksandr Razin$^{*}$\\
    {\tt\small razin.ad@edu.spbstu.ru}
\end{minipage}\hfill
\begin{minipage}[t]{0.3\textwidth}
    \centering
    Kazantsev Danil$^{*}$\\
    {\tt\small 311495@niuitmo.ru}
\end{minipage}\hfill
\begin{minipage}[t]{0.3\textwidth}
    \centering
    Ilya Makarov\\
    {\tt\small iamakarov@hse.ru}
\end{minipage}
}

\begin{document}
\maketitle
\blfootnote{$^{*}$ Equal Contribution.}

\begin{abstract}
Diffusion models struggle to scale beyond their training resolutions, as direct high-resolution sampling is slow and costly, while post-hoc image super-resolution (ISR) introduces artifacts and additional latency by operating after decoding. We present the Latent Upscaler Adapter (LUA), a lightweight module that performs super-resolution directly on the generator’s latent code before the final VAE decoding step. LUA integrates as a drop-in component, requiring no modifications to the base model or additional diffusion stages, and enables high-resolution synthesis through a single feed-forward pass in latent space. A shared Swin-style backbone with scale-specific pixel-shuffle heads supports $\times2$ and $\times4$ factors and remains compatible with image-space SR baselines, achieving comparable perceptual quality with nearly 3$\times$ lower decoding and upscaling time (adding only +0.42\,s for $1024$px generation from $512$px, compared to 1.87\,s for pixel-space SR using the same SwinIR architecture). Furthermore, LUA shows strong generalization across the latent spaces of different VAEs, making it easy to deploy without retraining from scratch for each new decoder. Extensive experiments demonstrate that LUA closely matches the fidelity of native high-resolution generation while offering a practical and efficient path to scalable, high-fidelity image synthesis in modern diffusion pipelines. Code: \rurl{github.com/vaskers5/LUA}.
\end{abstract}
\section{Introduction}
\label{sec:intro}

Diffusion models have transformed image synthesis, progressing from pixel-space formulations~\cite{ho2020denoising, song2020denoising} to \textit{Latent Diffusion Models} (LDMs), which shift computation into compact latent representations~\cite{rombach2022high}. This latent formulation underpins contemporary systems for image generation, editing, and translation~\cite{esser2024scaling, batifol2025flux}. Despite these advances, current models are effectively constrained by the spatial resolutions seen during training (typically $512^2$ or $1024^2$). Naïvely sampling beyond these scales often yields repetition, geometric distortions, and texture breakdown~\cite{he2023scalecrafter, zhang2024hidiffusion}. While retraining or high-resolution fine-tuning can reduce such artifacts~\cite{li2024scalability}, these remedies demand substantial compute and data.

\begin{figure}[t]
    \centering
    \includegraphics[width=\linewidth]{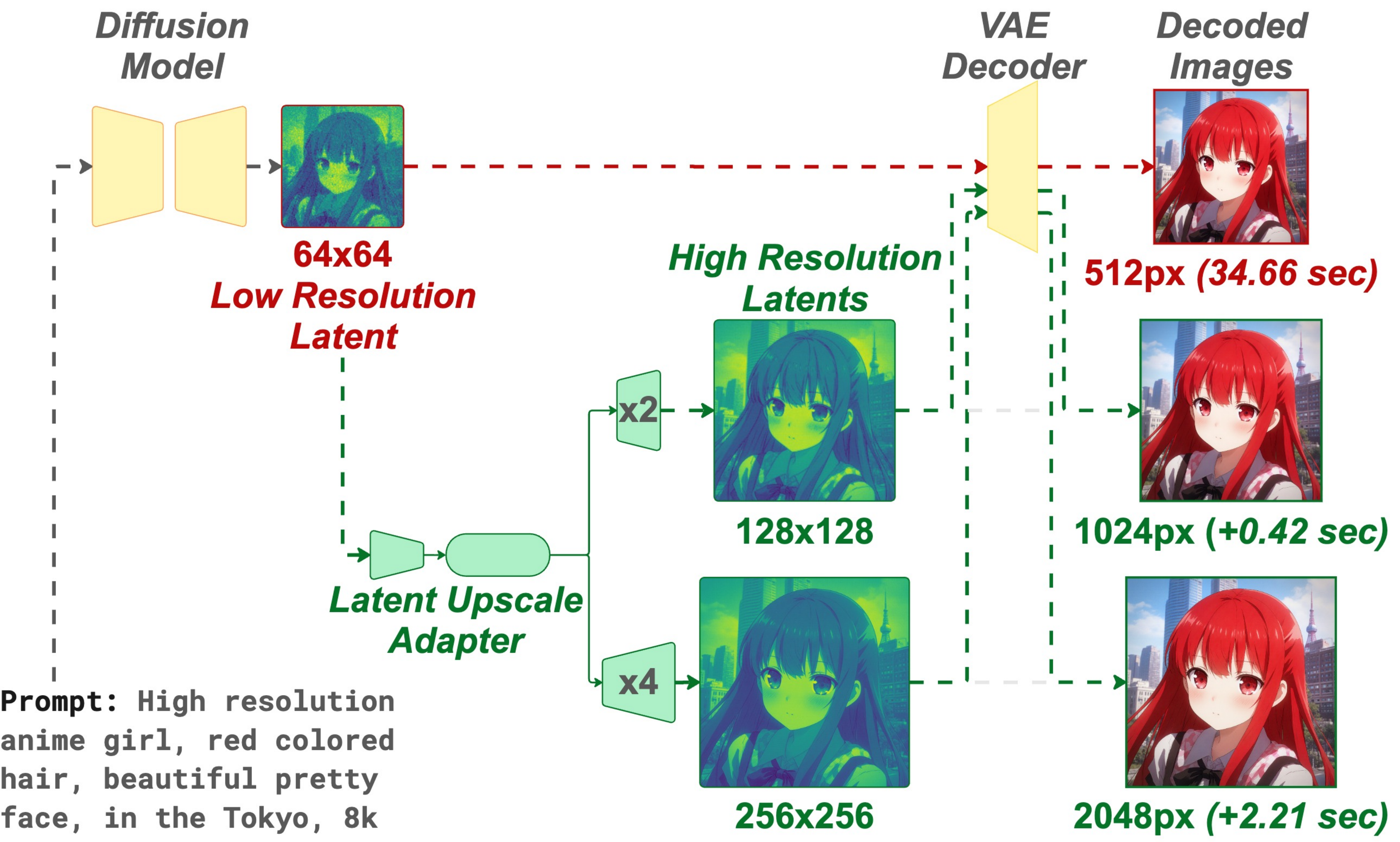}
    \caption{
        Our proposed lightweight \emph{Latent Upscaler Adapter} (LUA) integrates into diffusion pipelines \emph{without} retraining the generator/decoder and \emph{without} an extra diffusion stage. The example uses a FLUX~\cite{batifol2025flux} generator: it produces a $64{\times}64$ latent for a $512$\,px image (red dashed path decodes directly). Our path (green dashed) upsamples the same latent to $128{\times}128$ ($\times2$) or $256{\times}256$ ($\times4$) and decodes once to $1024$\,px or $2048$\,px, adding only $+0.42$\,s (1K) and $+2.21$\,s (2K) on an NVIDIA L40S GPU. LUA outperforms multi-stage high-resolution pipelines while avoiding their extra diffusion passes, and achieves efficiency competitive with image-space SR at comparable perceptual quality, all via a single final decode.
    }
    \label{fig:teaser}
\end{figure}

A pragmatic alternative is to generate at the native resolution and subsequently upsample. Two main paradigms implement this strategy. \textit{Pixel-space super-resolution} (SR) applies an external SR model to the decoded image. This approach is simple but reconstructs fine structure purely from pixels, which encourages oversmoothing, semantic drift, and a computational cost that grows quadratically with output size~\cite{chen2021hierarchical, wu2024seesrsemanticsawarerealworldimage}. \textit{Latent-space upsampling} instead enlarges the latent representation prior to decoding, reducing inference cost and better preserving semantics. Recent reference-based methods, such as DemoFusion-style pipelines~\cite{du2024demofusion} and LSRNA~\cite{jeong2025latent}, first generate a low-resolution reference, upsample it, and then run a second diffusion stage guided by the upsampled latent. Although effective, these pipelines require multi-stage inference, auxiliary noise or guidance branches, and tight coupling to specific VAEs, which increases latency and limits generality across model families.

\begin{figure}[t]
    \centering
    \includegraphics[width=\linewidth]{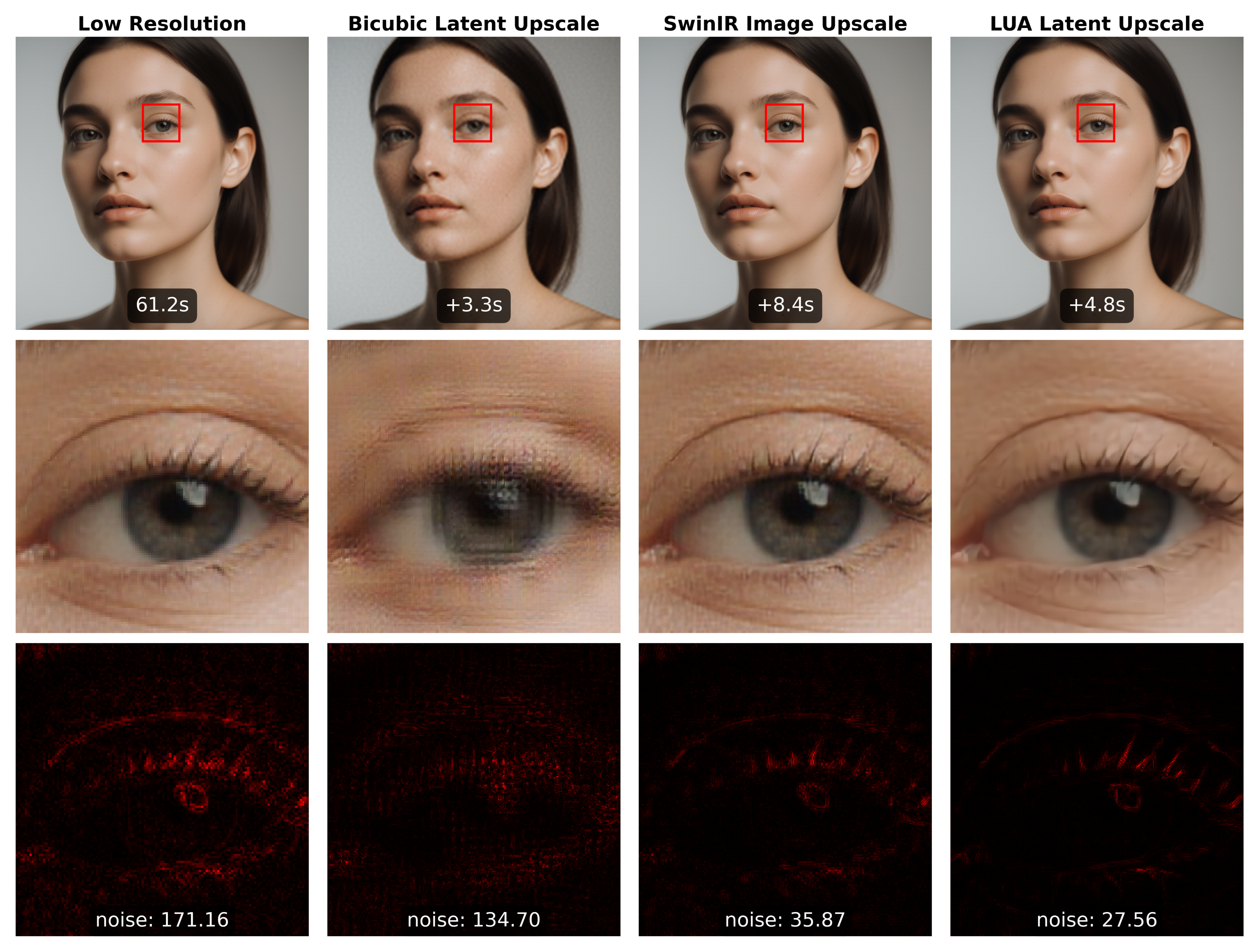}
    \caption{Upscaling FLUX outputs~\cite{batifol2025flux} from $1024^2{\rightarrow}2048^2$. Columns: (1) base decode, (2) bicubic \emph{latent}, (3) SwinIR \emph{image}-space SR, (4) LUA \emph{latent}-space SR. Top: runtime overhead vs.\ (1). Middle ($8\times$ crops): bicubic blurs/aliases; SwinIR sharpens but adds noise/texture drift; LUA preserves eyelashes and skin with stable edges. Bottom: Laplacian-variance maps (darker\,{=}\,less noise) with means—LUA attains the lowest residual noise and the smallest overhead via single-decode latent upscaling.}
    \label{fig:methods-compare-variance}
\end{figure}

In practice, the \textit{upsampling step} is the principal bottleneck. Bicubic resizing or naïve latent interpolation departs from the manifold of valid latents, producing unnatural textures after decoding (Fig.~\ref{fig:methods-compare-variance}, column 2). Conversely, pixel-space SR applied after decoding can improve fidelity but incurs higher latency and may introduce noise (Fig.~\ref{fig:methods-compare-variance}, column 3). The core challenge is to increase latent resolution while preserving manifold geometry and high-frequency latent details that decode into realistic images—\emph{without} invoking an additional diffusion process.

We address this challenge with the \textbf{Latent Upscaler Adapter (LUA)}, a lightweight module inserted between the generator and VAE decoder. Given a latent $z \in \mathbb{R}^{h\times w\times C}$, LUA predicts an upscaled latent $\hat{z} \in \mathbb{R}^{\alpha h\times \alpha w\times C}$ for $\alpha \in {2,4}$, which is decoded once to produce the final image. Because VAE decoders typically expand spatial dimensions by stride $s{=}8$, a $\times2$ latent upscaling yields a $16\times$ increase in pixel count with no extra denoising. LUA employs a shared SwinIR-style backbone~\cite{liang2021swinir} with lightweight, scale-specific pixel-shuffle heads and is trained directly in the latent space with complementary latent- and pixel-domain objectives to preserve high-frequency consistency and enable photorealistic decoding (Fig.~\ref{fig:methods-compare-variance}, column~4).

Our study focuses on three questions:
\begin{enumerate}
    \item Can a simple latent adapter deliver higher-quality images at lower cost than native high-resolution synthesis or pixel-space SR?
    \item Can a single adapter handle multiple upscaling factors within one unified framework?
    \item Can an adapter trained for one model’s VAE transfer to other generators with minimal fine-tuning?
\end{enumerate}

We evaluate LUA across backbones and resolutions using FID/KID/CLIP and local-detail metrics, with ablations on architecture, objectives, and scale handling. Preserving latent microstructure via a single upscaling stage narrows the gap to native high-res synthesis while reducing complexity and latency relative to multi-stage diffusion and pixel-space SR.

Our contributions are threefold:
\begin{itemize}
\item We train \emph{only} a latent upscaler adapter with a multi-stage curriculum and show that latent-space super-resolution attains quality comparable to modern high-resolution diffusion pipelines while being more efficient and less noisy than pixel-space SR.
\item We design a single model that supports multiple scale factors ($\times2$, $\times4$) via a shared backbone with jointly trained, scale-specific heads—avoiding retraining from scratch.
\item We demonstrate cross-VAE generalization: the same backbone operates across SD3~\cite{esser2024scaling}, SDXL~\cite{podell2023sdxlimprovinglatentdiffusion}, and FLUX~\cite{batifol2025flux} by changing only the first layer to match input channels with minimal fine-tuning.
\end{itemize}

\section{Related Work}
\label{sec:related_work}

This section reviews three lines of work relevant to high-resolution image synthesis: (i) efficient diffusion-based generation at large scales, (ii) super-resolution in pixel and latent spaces, and (iii) multi-scale (discrete vs.\ continuous) super-resolution. For each, we outline representative methods and the limitations that motivate our approach.

\paragraph{Efficient high-resolution generation with diffusion models.}
Diffusion models in compressed latent spaces~\cite{rombach2022high} enable controllable synthesis, yet sampling beyond the training scale ($512^2$–$1024^2$) often yields repetition, distortions, or texture loss~\cite{he2023scalecrafter, zhang2024hidiffusion}. Direct high-resolution training has been demonstrated in large systems such as SDXL~\cite{podell2023sdxlimprovinglatentdiffusion}, SD3~\cite{esser2024scaling}, and related frameworks~\cite{li2024scalability}, but it demands massive datasets and compute. Inference-time strategies avoid full retraining: tiling/blending in MultiDiffusion~\cite{bar2023multidiffusion} preserves locality but risks seams; receptive-field expansions via adaptive/dilated or sparse convolutions~\cite{fayyaz2022adaptive} and step-reduction by distillation/consistency methods~\cite{meng2023distillation, song2023consistency} improve speed but can reduce fidelity at extreme resolutions; progressive pipelines such as HiDiffusion and ScaleCrafter~\cite{zhang2024hidiffusion, he2023scalecrafter} iteratively upsample and refine; reference-based approaches like DemoFusion~\cite{du2024demofusion} synthesize a low-resolution latent, upsample it, and re-diffuse. These families rely on additional diffusion passes with method-specific schedules and step counts, increasing latency and coupling execution to particular resolution settings.

\paragraph{Super-resolution in image and latent spaces.}
Pixel-space SR advanced from CNN regressors (SRCNN, EDSR)~\cite{dong2015image, lim2017enhanceddeepresidualnetworks} to adversarial/perceptual (SRGAN, ESRGAN)~\cite{ledig2017photorealisticsingleimagesuperresolution, wang2018esrganenhancedsuperresolutiongenerative} and transformer models (SwinIR, HAT)~\cite{liang2021swinir, chen2025hathybridattentiontransformer}; diffusion-based SR (SR3, SRDiff, SeeSR, StableSR, DiffBIR, SUPIR)~\cite{saharia2021imagesuperresolutioniterativerefinement, li2021srdiffsingleimagesuperresolution, wu2024seesrsemanticsawarerealworldimage, wang2024exploitingdiffusionpriorrealworld, lin2024diffbirblindimagerestoration, yu2024scalingexcellencepracticingmodel} further boosts perceptual fidelity. However, all denoise at the target resolution, incurring quadratic compute/memory with image size and risking semantic drift in fine textures. Latent-space SR reduces cost by upsampling before decoding, but naïve latent interpolation (e.g., bicubic/linear) departs from the generative manifold, and learned mappings such as LSRNA or the latent guidance used in DemoFusion~\cite{jeong2025latent, du2024demofusion} still depend on a subsequent diffusion stage, yielding multi-stage, latency-heavy pipelines.

\paragraph{Discrete vs.\ continuous multi-scale SR.}
Discrete-factor SR (e.g., $\times2$, $\times4$) commonly trains separate networks or employs a shared backbone with scale-specific heads (MDSR, SwinIR)~\cite{lim2017enhanceddeepresidualnetworks, liang2021swinir}, which is effective but resource intensive to train and store for multiple scales. Arbitrary-scale methods (LIIF, LTE, CiaoSR)~\cite{chen2021learningcontinuousimagerepresentation, lee2022localtextureestimatorimplicit, cao2023ciaosrcontinuousimplicitattentioninattention} predict continuous coordinates from learned features, yet often underperform direct upscaling on fine textures where high-frequency structures dominate.

In contrast, we target a single-pass \emph{latent} upscaler that avoids extra diffusion stages, side-steps the quadratic cost and drift of pixel-space SR, and replaces naïve latent resizing with a dedicated training curriculum. The design supports multiple scales via a shared backbone with lightweight, scale-specific heads rather than separate per-scale models or weaker arbitrary-scale decoders. This aims to deliver high-resolution fidelity with substantially lower latency and practical deployment characteristics.

\begin{figure}[t]
    \centering
    \includegraphics[width=\linewidth]{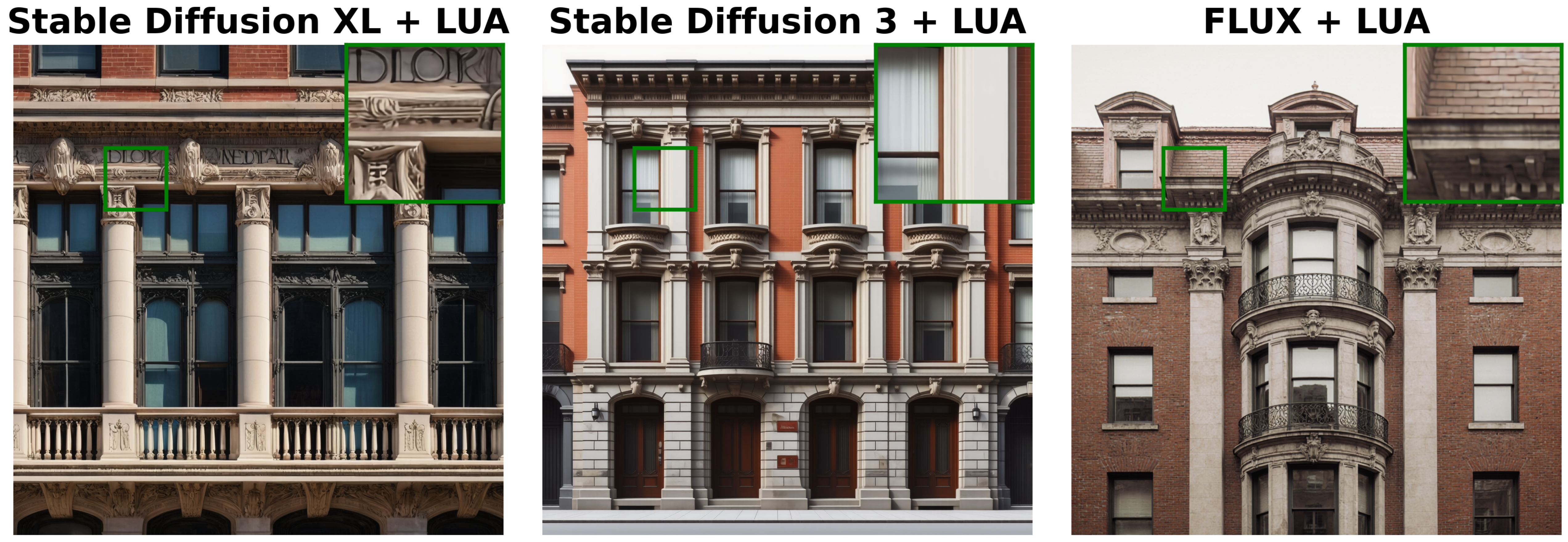}
    \caption{Cross-model $2\times$ latent upscaling with a single adapter. For SDXL~\cite{podell2023sdxlimprovinglatentdiffusion}, SD3~\cite{esser2024scaling}, and FLUX~\cite{batifol2025flux}, a $128{\times}128$ latent is upscaled to $256{\times}256$ by the same LUA and decoded once by each model’s native VAE to yield $2048^2$ images. SD3 and FLUX share $C{=}16$ latents; SDXL ($C{=}4$) is supported by changing only the first convolution. Insets show artifact-free detail preservation; green boxes mark $\times8$ zooms.}
    \label{fig:cross_model_examples}
\end{figure}

\section{Proposed Method}
\label{sec:method}

We target high-resolution synthesis without an extra diffusion stage or retraining the generator/VAE by inserting a single-pass \emph{Latent Upscaler Adapter} (LUA) between the generator and the frozen decoder: LUA enlarges the latent, then one decode produces the final image with near–base-decode overhead. In Sec.~\ref{sec:latent_upscaling} we formalize the upscaling operator $U_\alpha$, establish its computational efficiency relative to image-space SR and multi-stage diffusion, and show cross-model generalization (FLUX, SD3, SDXL) via changing only the first convolution layer and minimal fine-tuning. We then describe the multi-scale \emph{architecture}—a shared backbone with scale-specific heads for $\times2$ and $\times4$ (Sec.~\ref{architecture})—and the \emph{multi-stage training} strategy that preserves latent microstructure and stabilizes decoded appearance (Sec.~\ref{Multi-stage-train}).

\subsection{Latent Upscaling}
\label{sec:latent_upscaling}

\paragraph{Formulation.}
Given text condition $c$ and noise $\epsilon$, a pretrained generator $G$ produces a latent $z\in\mathbb{R}^{h\times w\times C}$:
\begin{equation}
z = G(c, \epsilon).
\end{equation}
A frozen VAE decoder $D$ with spatial stride $s$ (typically $s{=}8$) maps $z$ to an RGB image $x \in \mathbb{R}^{(s h) \times (s w) \times 3}$:
\begin{equation}
x = D(z).
\end{equation}

We introduce a deterministic latent upscaler $U_{\alpha}$ with scale factor $\alpha\!\in\!\{2,4\}$ that maps $z\!\in\!\mathbb{R}^{h\times w\times C}$ to $\hat{z}\!\in\!\mathbb{R}^{\alpha h\times \alpha w\times C}$:
\begin{equation}
\hat{z} = U_{\alpha}(z).
\end{equation}
A single decode yields the high-resolution image:
\begin{equation}
\hat{x} = D(\hat{z}).
\end{equation}
Here, $h,w$ are the latent spatial dimensions, $C$ is the latent channel width, $s$ is the decoder stride, $c$ is the conditioning (e.g., text embedding), and $\epsilon$ is the generator noise. All generative stochasticity resides in $G$; $U_{\alpha}$ is a feed-forward operator trained to remain on the latent manifold and to preserve the fine-scale statistics required for photorealistic decoding.

\paragraph{Computational efficiency.}
Pixel-space SR operates on $(sh)\!\times\!(sw)$ positions; LUA operates on $h\!\times\!w$ positions and still decodes \emph{once}. The cost ratio scales as
\begin{equation}
\frac{O((sh)(sw))}{O(hw)} \approx s^2,
\end{equation}
so for typical $s{=}8$ our upscaler touches about $1/64$ as many spatial elements as image-space SR. In addition, unlike progressive or reference-based pipelines, LUA does not add a second diffusion pass or any full-resolution refinement stage. Empirically, this yields lower wall-clock overhead and memory traffic while achieving comparable high-resolution fidelity (Sec.~\ref{sec:experiments}).

\paragraph{Cross-model generalization.}
LUA acts on latents, not backbone internals, and therefore does not require training from scratch for each VAE. The same model generalizes across FLUX, SD3, and SDXL by changing only the \emph{first convolution layer} to match the input channel count and fine-tuning the adapter on a small set of latents from the target model. The backbone and scale heads remain unchanged. Architectural details are given in Sec.~\ref{architecture}; the fine-tuning protocol and data sizes are in Sec.~\ref{sec:experiments}.

\begin{figure}[t]
    \centering
    \includegraphics[width=\linewidth]{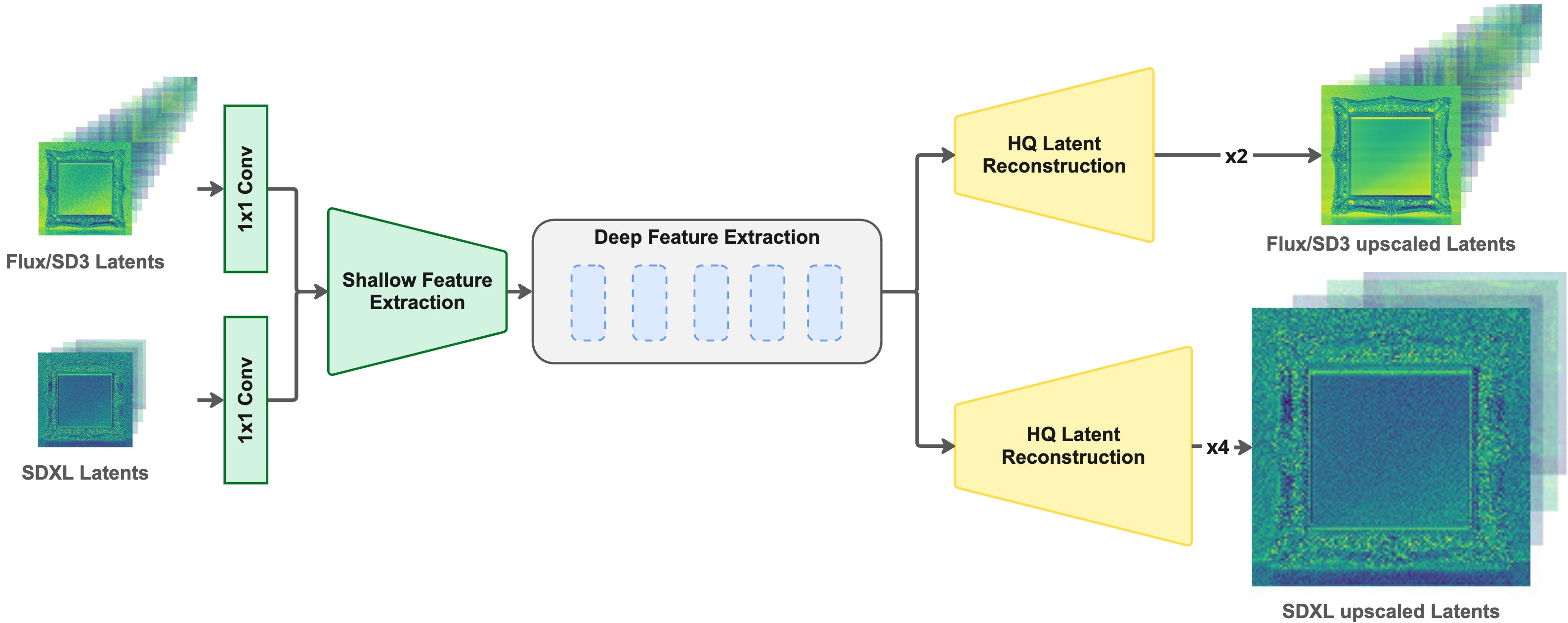}
    \caption{Architecture of the Latent Upscaler Adapter (LUA). A SwinIR-style backbone~\cite{liang2021swinir} is shared across scales; a $1{\times}1$ input conv adapts the VAE latent width ($C{=}16$ for FLUX/SD3; $C{=}4$ for SDXL). Scale-specific pixel-shuffle heads output $\times2$ or $\times4$ latents. At inference, the path selects the input adapter, runs the shared backbone, and activates the requested head. The schematic shows FLUX/SD3 $\times2$ and SDXL $\times4$.}
    \label{fig:LUA}
\end{figure}

\subsection{Architecture}
\label{architecture}

We adopt a Swin Transformer restoration backbone in the spirit of SwinIR~\cite{liang2021swinir}, which has proven effective for super-resolution in RGB space, and consistent with latent-domain adaptations such as LSRNA~\cite{jeong2025latent}. Windowed self-attention provides long-range context while preserving locality, matching the spatial–statistical structure of VAE latents. Given an input latent $z \in \mathbb{R}^{h \times w \times C}$, the backbone $\phi(\cdot)$ extracts features through an encoder–decoder with residual connections. Upscaling is realized with explicit SR heads—shallow convolutions followed by pixel-shuffle—that directly predict an enlarged latent; compared with implicit coordinate decoders (e.g., LIIF~\cite{chen2021learningcontinuousimagerepresentation}), this better preserves high-frequency latent microstructure that decodes into sharper textures.

To support multiple scale factors without duplicating capacity, a single backbone $\phi$ is shared across scale-specific heads for $\times2$ and $\times4$ (denoted $U_{\times2}$ and $U_{\times4}$). Joint training with balanced sampling encourages the backbone to learn scale-agnostic representations while each head specializes to its factor’s aliasing and artifact profile. At inference, the backbone runs once and the head corresponding to the requested scale $\alpha \in \{2,4\}$ is applied:
\begin{equation}
\hat{x} = D\!\big(U_{\times\alpha}(\phi(z))\big), \qquad \alpha \in \{2,4\},
\end{equation}
where $D$ is the frozen VAE decoder and $z$ is the input latent with channel width $C$. The overall module layout is depicted in Fig.~\ref{fig:LUA}.

To operate across VAEs with different latent channel widths (e.g., SD3/FLUX with $C{=}16$ and SDXL with $C{=}4$), we change only the first convolution layer to match the input channels; the shared backbone and the scale heads are reused unchanged. A brief fine-tuning on a small set of latents from the target model aligns statistics and enables cross-VAE transfer without training from scratch. The fine-tuning protocol and data sizes are detailed in Sec.~\ref{sec:experiments}, and cross-model results are shown in Fig.~\ref{fig:cross_model_examples}.

\subsection{Multi-stage Training Strategy}
\label{Multi-stage-train}

Single-domain objectives are insufficient for latent SR: optimizing \emph{only} in latent space preserves coarse structure but yields decoded images with residual grid/blur and spurious high-frequency noise, while optimizing \emph{only} in pixel space is unstable—gradients backpropagated through the frozen VAE decoder interact with unnormalized latents and fail to converge. To address both issues, we adopt a progressive three-stage curriculum that first secures latent structural and spectral alignment, then couples latent fidelity to decoded appearance with high-frequency emphasis, and finally refines edges in pixel space without re-diffusion. Throughout this section, $z$ and $\hat{z}$ denote input and upscaled latents, $z_{\mathrm{HR}}$ the reference HR latent (from the frozen VAE encoder), and $x{=}D(z)$, $\hat{x}{=}D(\hat{z})$, $x_{\mathrm{HR}}{=}D(z_{\mathrm{HR}})$ their decoded images via the frozen decoder $D$. Superscripts ${}^{z}$ and ${}^{x}$ indicate losses computed in latent and pixel domains, respectively.

\textit{Stage I — Latent-domain structural alignment.}
This stage learns a stable mapping $\hat{z}=U_\alpha(z)$ that matches high-resolution latent structure and spectra while avoiding over-smoothing. The objective is
\begin{equation}
\mathcal{L}_{\mathrm{SI}}
= \alpha_1 \, \mathcal{L}_{\mathrm{L1}}^{z}
+ \beta_1 \, \mathcal{L}_{\mathrm{FFT}}^{z},
\end{equation}
where $\alpha_1,\beta_1 \ge 0$ balance reconstruction and spectral alignment in latent space, with
\begin{align}
\mathcal{L}_{\mathrm{L1}}^{z} &= \big\lvert \hat{z} - z_{\mathrm{HR}} \big\rvert_{1}, \\
\mathcal{L}_{\mathrm{FFT}}^{z} &= \big\lvert \mathcal{F}(\hat{z}) - \mathcal{F}(z_{\mathrm{HR}}) \big\rvert_{1}.
\end{align}
Here $\hat{z}\in\mathbb{R}^{\alpha h \times \alpha w \times C}$ and $\mathcal{F}(\cdot)$ is the channel-wise 2D FFT magnitude~\cite{fuoli2021fourierspacelossesefficient}. $\mathcal{L}_{\mathrm{L1}}^{z}$ enforces element-wise correspondence; $\mathcal{L}_{\mathrm{FFT}}^{z}$ aligns high-frequency latent statistics to preserve microstructure~\cite{jiang2024fast}.

\textit{Stage II — Joint latent–pixel consistency.}
This stage links latent fidelity to decoded appearance by adding image-domain constraints while retaining the Stage~I latent terms:
\begin{equation}
\mathcal{L}_{\mathrm{SII}}
= \alpha_2 \, \mathcal{L}_{\mathrm{L1}}^{z}
+ \beta_2 \, \mathcal{L}_{\mathrm{FFT}}^{z}
+ \gamma_2 \, \mathcal{L}_{\mathrm{DS}}^{x}
+ \delta_2 \, \mathcal{L}_{\mathrm{HF}}^{x},
\end{equation}
with $\alpha_2,\beta_2,\gamma_2,\delta_2 \ge 0$ and
\begin{align}
\mathcal{L}_{\mathrm{DS}}^{x} &=
\big\lvert \downarrow_d(\hat{x}) - \downarrow_d(x_{\mathrm{HR}}) \big\rvert_{1}, \\
\mathcal{L}_{\mathrm{HF}}^{x} &=
\Big\lvert
\big(\hat{x} - \mathcal{G}_\sigma(\hat{x})\big)
-
\big(x_{\mathrm{HR}} - \mathcal{G}_\sigma(x_{\mathrm{HR}})\big)
\Big\rvert_{1}.
\end{align}

Here $\downarrow_{d}(\cdot)$ denotes bicubic downsampling~\cite{rad2020benefitingbicubicallydownsampledimages} with $d{=}2$ for $\times2$ and $d{=}4$ for $\times4$, and $\mathcal{G}_\sigma(\cdot)$ is a Gaussian blur with $\sigma{=}1.0$ (applied channel-wise). $\mathcal{L}_{\mathrm{DS}}^{x}$ enforces coarse appearance consistency at a common reduced scale; $\mathcal{L}_{\mathrm{HF}}^{x}$ emphasizes edges and textures by matching high-frequency residuals.

\textit{Stage III — Edge-aware image refinement.}
The final stage sharpens edges and suppresses residual ringing/grids in pixel space, without any additional denoising:
\begin{equation}
\mathcal{L}_{\mathrm{SIII}}
= \alpha_3 \, \mathcal{L}_{\mathrm{L1}}^{x}
+ \beta_3 \, \mathcal{L}_{\mathrm{FFT}}^{x}
+ \gamma_3 \, \mathcal{L}_{\mathrm{EAGLE}}^{x},
\end{equation}
with $\alpha_3,\beta_3,\gamma_3\ge 0$ and
\begin{align}
\mathcal{L}_{\mathrm{L1}}^{x} &= \big\lvert \hat{x} - x_{\mathrm{HR}} \big\rvert_{1}, \\
\mathcal{L}_{\mathrm{FFT}}^{x} &= \big\lvert \mathcal{F}(\hat{x}) - \mathcal{F}(x_{\mathrm{HR}}) \big\rvert_{1},
\end{align}
where $\mathcal{L}_{\mathrm{EAGLE}}^{x}$ is an edge-aware gradient localization loss that enforces crisp boundaries and reduces staircase artifacts~\cite{sun2025eagle}.

\noindent\textit{Weight selection.}
Weights were set via grid search: $\ell_1$ and FFT terms (latent and image domains) receive the largest coefficients as primary reconstruction and texture-sensitive objectives, while downsampling and edge-aware terms use smaller auxiliary regularizers. Please see the appendix for detailed grid-search settings and per-stage loss weights.

\begin{figure}[t]
    \centering
    \includegraphics[width=\linewidth]{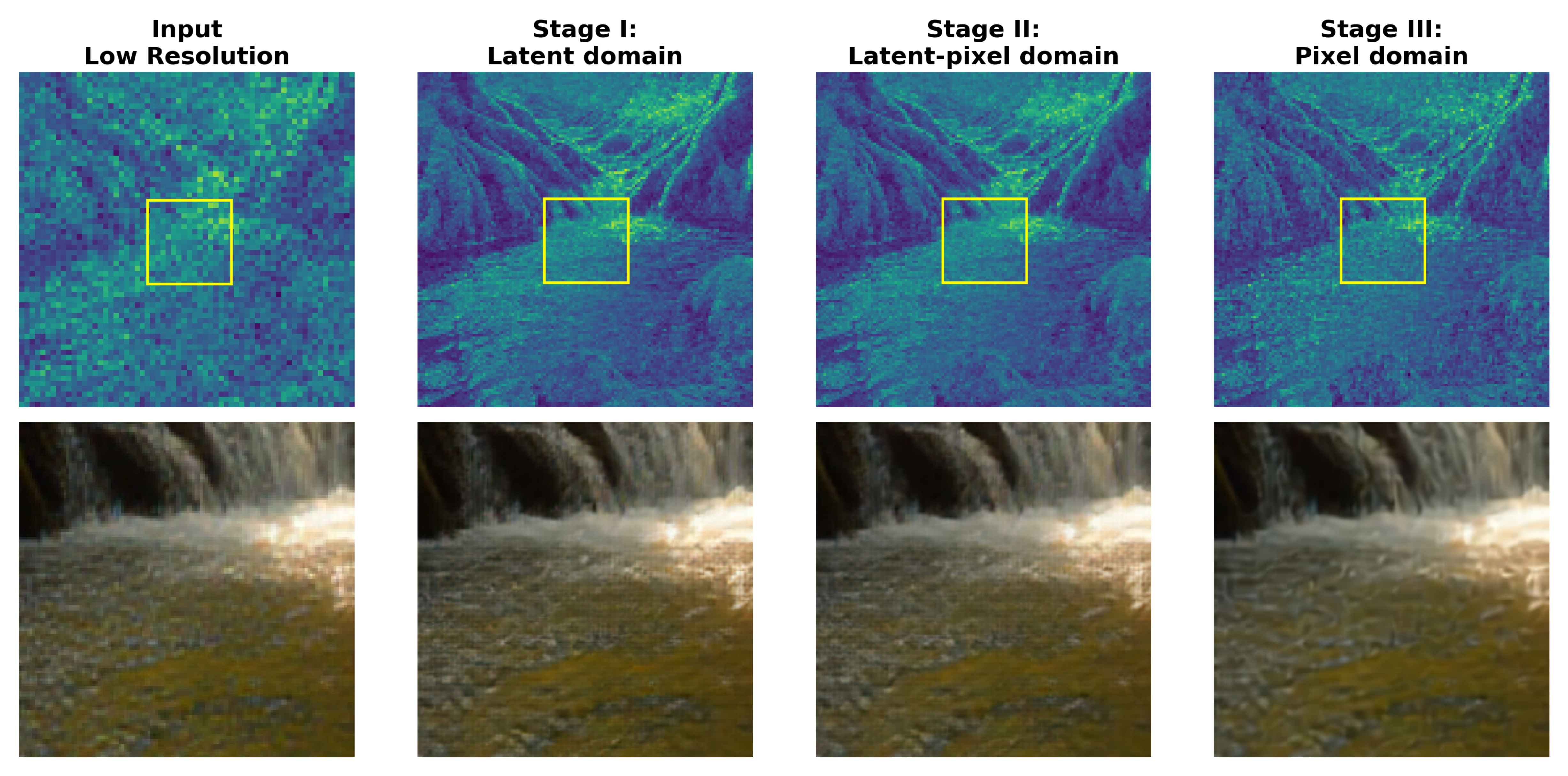}
    \caption{Effect of the three-stage curriculum on latent reconstruction and decoded appearance (FLUX backbone). The $2{\times}4$ grid shows \emph{top}: latent feature maps (channel 10, min–max normalized); \emph{bottom}: corresponding $8{\times}$ zoomed decodes. Columns: (1) original low-resolution latent ($128^2$) and decode; (2–4) LUA upscaled latents to $256^2$ after Stage~I–III with their decodes. Yellow boxes mark the zoomed region. From (2) to (4), decodes become less noisy and more structured; Stage~III concentrates high-frequency energy around details, indicating that controlled latent noise aids faithful VAE decoding.}
    \label{fig:multi_stage_zoom}
\end{figure}

Our multi-stage latent–pixel curriculum adapts LUA to the latent domain and the frozen decoder. Stage~I aligns the upscaled latent with high-resolution structure and spectra (Fig.~\ref{fig:multi_stage_zoom}, col.~2). Stage~II introduces joint latent–image supervision (downsampled and high-frequency terms) to normalize latents before decoding and suppress noise (col.~3). Stage~III trains in pixel space only, selectively removing redundant noise and grid artifacts while preserving necessary high-frequency detail (col.~4). The curriculum enables coherent cross-domain training and yields single-decode synthesis without an extra diffusion stage; per-stage loss weights are provided in the appendix.
\section{Experiments}
\label{sec:experiments}

We evaluate LUA on OpenImages with a unified train/val setup, comparing to representative high-resolution baselines and reporting fidelity (FID/KID and patch variants), text–image alignment (CLIP), and wall-clock latency. We also study cross-model and multi-scale generalization and ablate the curriculum and architecture. Further implementation details and extended results are provided in the appendix.

\subsection{Experiment Settings}
\label{sec:exp_settings}

\textbf{Training data.}
Following LSRNA~\cite{jeong2025latent}, we use OpenImages~\cite{Kuznetsova_2020}; photos with both sides $\geq 1440$\,px are tiled into non-overlapping $512{\times}512$ crops. HR/LR pairs are made by bicubic downsampling at the target scale ($\times2$, $\times4$). Crops are encoded with the FLUX VAE (stride $s{=}8$, $C{=}16$) to produce latent: for $\times2$, $16{\times}32{\times}32 \rightarrow 16{\times}64{\times}64$; for $\times4$, $16{\times}16{\times}16 \rightarrow 16{\times}64{\times}64$. The final set contains 3.8M pairs.

\textbf{Multi-stage training.}
We train the three stages (Sec.~\ref{Multi-stage-train}) with Adam (lr $2{\times}10^{-4}$, weight decay $0$), EMA $0.999$, grad-clip $0.4$, and MultiStepLR (milestones $62.5$k/$93.75$k/$112.5$k, $\gamma{=}0.5$); each stage runs $125$k steps. Hyperparameters follow SwinIR~\cite{liang2021swinir}. We keep the same initial lr across stages; Stage~III uses a short warm-up. Effective batch sizes: $2{,}048$ (Stage~I) and $32$ (Stages~II–III). Losses match Sec.~\ref{Multi-stage-train}.

\textbf{Scalability: multi-scale and cross-VAE.}
\label{multi-scale-protocol}
A single Swin backbone is shared by the $\times2/\times4$ heads ($U_{\times2}$, $U_{\times4}$) and trained jointly with balanced sampling (Sec.~\ref{sec:method}, Fig.~\ref{fig:LUA}). For cross-VAE transfer, we build 500k LR/HR latent pairs for SDXL ($C{=}4$) and SD3 ($C{=}16$), replace only the first convolution to match channels, and perform a brief Stage~III–style fine-tune, reusing the backbone and heads without retraining from scratch.

\subsection{Evaluation Protocol}
\label{sec:eval_protocol}

\textbf{Validation data.}
We evaluate on 1{,}000 held-out high-resolution OpenImages photos (disjoint from training). Prompts are obtained via captioning and shared across methods; see the appendix for details.

\textbf{Metrics.}
We report FID~\cite{heusel2018ganstrainedtimescaleupdate}, KID~\cite{bińkowski2021demystifyingmmdgans}, and CLIP~\cite{radford2021learningtransferablevisualmodels}. Images are synthesized at the target resolution ($1024^2/2048^2/4096^2$). To better capture fine detail, we also report patch metrics (pFID/pKID) on random $1024{\times}1024$ crops with an identical crop policy across methods, following LSRNA and Anyres-GAN~\cite{chai2022anyresolutiontraininghighresolutionimage}. Additional metrics and computation details are provided in the appendix.

\textbf{Baselines.}
We compare ScaleCrafter~\cite{he2023scalecrafter}, HiDiffusion~\cite{zhang2024hidiffusion} (progressive), DemoFusion~\cite{du2024demofusion}, LSRNA–DemoFusion~\cite{jeong2025latent} (reference-based re-diffusion), SDXL (Direct), and SDXL+SwinIR~\cite{liang2021swinir} (pixel-space SR). Samplers, steps, guidance, and prompts are held constant where applicable; see the appendix for configuration details.

\textbf{Our method.}
\emph{SDXL+LUA}: generate at $1024^2$, upscale in latent space, decode once. We report SDXL-anchored comparisons in Table~\ref{tab:quantitative_results} and summarize cross-model/multi-scale results (FLUX/SD3/SDXL at $\times2/\times4$) in Table~\ref{tab:lua_cross_model_multiscale}.

\textbf{Runtime measurement.}
Per-image wall-clock time is the median over $N$ runs after 20 warm-ups on a single GPU with AMP and batch size~1. Composite pipelines include generation, upscaling, and decode. 

\begin{table*}[t]
  \centering
  \small
  \setlength{\tabcolsep}{5pt}%
  \renewcommand{\arraystretch}{1.05}%
  \setlength{\aboverulesep}{0pt}%
  \setlength{\belowrulesep}{0pt}%
  \setlength{\extrarowheight}{0pt}%
  \caption{OpenImages validation. Metrics follow Sec.~\ref{sec:eval_protocol} (FID/pFID, KID/pKID, CLIP) and runtime (median s). H100, batch size 1. For \textbf{1K} ($1024^2$), SDXL samples at $512^2$ then upscales to $1024^2$ via LUA (latent) or SwinIR (pixel). For \textbf{2K}/\textbf{4K}, images are sampled at $1024^2$ then upscaled to $2048^2$/$4096^2$. LUA achieves the lowest latency at all resolutions and the strongest fidelity at 2K/4K; best marked in \textbf{bold}.}
  \label{tab:quantitative_results}
  \begin{tabular*}{\textwidth}{@{\extracolsep{\fill}} l l cc cc c c}
    \toprule
    Resolution & Method & FID $\downarrow$ & pFID $\downarrow$ & KID $\downarrow$ & pKID $\downarrow$ & CLIP $\uparrow$ & Time (s) \\
    \midrule
    \multirow{6}{*}{$1024{\times}1024$}
      & HiDiffusion                  & 232.55 & 230.39 & 0.0211 & 0.0288 & 0.695 & 1.54 \\
      & DemoFusion                   & 195.82 & 193.99 & 0.0153 & 0.0229 & 0.725 & 2.04 \\
      & LSRNA\textendash DemoFusion  & 194.55 & 192.73 & \textbf{0.0151} & 0.0228 & 0.734 & 3.09 \\
      & SDXL (Direct)                & \textbf{194.53} & 192.71 & \textbf{0.0151} & \textbf{0.0225} & 0.731 & 1.61 \\
      & SDXL + SwinIR                & 210.40 & 204.23 & 0.0313 & 0.0411 & 0.694 & 2.47 \\
      & \textbf{SDXL + LUA (ours)}   & 209.80 & \textbf{191.75} & 0.0330 & 0.0426 & \textbf{0.738} & \textbf{1.42} \\
    \midrule
    \multirow{6}{*}{$2048{\times}2048$}
      & HiDiffusion                  & 200.72 & 114.30 & 0.0030 & 0.0090 & 0.738 & 4.97 \\
      & DemoFusion                   & 184.79 & 177.67 & 0.0030 & 0.0100 & 0.750 & 28.99 \\
      & LSRNA\textendash DemoFusion  & 181.24 & 98.09 & 0.0019 & 0.0066 & 0.762 & 20.77 \\
      & SDXL (Direct)                & 202.87 & 116.57 & 0.0030 & 0.0086 & 0.741 & 7.23 \\
      & SDXL + SwinIR                & 183.16 & 100.09 & 0.0020 & 0.0077 & 0.757 & 6.29 \\
      & \textbf{SDXL + LUA (ours)}   & \textbf{180.80} & \textbf{97.90} & \textbf{0.0018} & \textbf{0.0065} & \textbf{0.764} & \textbf{3.52} \\
    \midrule
    \multirow{6}{*}{$4096{\times}4096$}
      & HiDiffusion                  & 233.65 & 95.95  & 0.0158 & 0.0214 & 0.698 & 122.62 \\
      & DemoFusion                   & 185.36 & 177.89 & 0.0043 & 0.0113 & 0.749 & 225.77 \\
      & LSRNA\textendash DemoFusion  & 177.95 & 62.07  & 0.0023 & \textbf{0.0071} & 0.757 & 91.64 \\
      & SDXL (Direct)                & 280.42 & 101.89 & 0.0396 & 0.0175 & 0.663 & 148.71 \\
      & SDXL + SwinIR                & 183.15 & 65.71  & 0.0018 & 0.0103 & 0.756 & 7.29 \\
      & \textbf{SDXL + LUA (ours)}   & \textbf{176.90} & \textbf{61.80}  & \textbf{0.0015} & 0.0152 & \textbf{0.759} & \textbf{6.87} \\
    \bottomrule
  \end{tabular*}
\end{table*}

\subsection{Quantitative Results}
\label{sec:quant_results}

Table~\ref{tab:quantitative_results} summarizes fidelity–efficiency trade-offs across resolutions. At $1024^2$, SDXL+LUA attains the lowest latency (1.42\,s) but lags strongest single-stage baselines on fidelity (FID 209.80 vs.\ 194.53 for SDXL~(Direct) and 194.55 for LSRNA–DemoFusion). We attribute this gap to low-resolution input latent ($64{\times}64$ for $512$\,px, stride~8), which constrains recoverable detail in $\times2$ setting; notably, our patch fidelity is competitive (pFID 191.75, best in row), indicating preserved local structure despite lower scores.

Beyond $1$K, LUA consistently improves both quality and speed. At $2048^2$, SDXL+LUA delivers the best fidelity among single-decode pipelines while remaining the fastest: FID 180.80, pFID 97.90, KID 0.0018, CLIP 0.764 in 3.52\,s, outperforming SDXL+SwinIR (183.16 / 100.09 / 0.0020 / 0.757 in 6.29\,s) and substantially undercutting multi-stage re-diffusion (LSRNA–DemoFusion: FID 181.24 in 20.77\,s; DemoFusion: 184.79 in 28.99\,s). At $4096^2$, SDXL+LUA again sets the best single-pass fidelity (FID 176.90; pFID 61.80) with the lowest runtime (6.87\,s), surpassing SDXL+SwinIR (183.15; 65.71; 7.29\,s) and avoiding the quality collapse of direct high-res sampling (SDXL~(Direct) FID 280.42). LSRNA–DemoFusion attains slightly lower pKID (0.0071) but is an order of magnitude slower (91.64\,s), underscoring LUA’s favorable accuracy–latency frontier. These results highlight the efficacy of LUA for high-resolution synthesis.

Table~\ref{tab:lua_cross_model_multiscale} further shows that a single backbone generalizes across models and scales with minimal adaptation: at $\times2$, FLUX+LUA reaches FID 180.99 and CLIP 0.773; at $\times4$, SDXL+LUA yields KID 0.0015 while FLUX+LUA attains the strongest pFID (62.30). These results confirm robust transfer across SDXL, SD3, and FLUX and consistent gains at both $\times2$ and $\times4$ without retraining the generator or adding diffusion stages. These findings highlight the adaptability of LUA across models and scales.

\begin{table*}[t]
  \centering
  \small
  \setlength{\tabcolsep}{5pt}%
  \renewcommand{\arraystretch}{1.05}%
  \setlength{\aboverulesep}{0pt}%
  \setlength{\belowrulesep}{0pt}%
  \setlength{\extrarowheight}{0pt}%
  \caption{Cross-model, multi-scale results for LUA. Metrics, crop protocol, runtimes, and hardware match Table~\ref{tab:quantitative_results}. We evaluate $\times2/\times4$ latent upscaling ($1024^2{\rightarrow}2048^2$/$4096^2$) on FLUX, SDXL, and SD3; best results are in \textbf{bold}.}
  \label{tab:lua_cross_model_multiscale}
  \begin{tabular*}{\textwidth}{@{\extracolsep{\fill}} l l cc cc c c}
    \toprule
    Scale & Diffusion Model & FID $\downarrow$ & pFID $\downarrow$ & KID $\downarrow$ & pKID $\downarrow$ & CLIP $\uparrow$ & Time (s) \\
    \midrule
    \multirow{3}{*}{$\times2$}
      & FLUX + LUA & \textbf{180.99} & \textbf{100.40} & 0.0020 & \textbf{0.0079} & \textbf{0.773} & 29.829 \\
      & SDXL + LUA & 183.15          & 101.18          & \textbf{0.0020} & 0.0087          & 0.753 & \textbf{3.52} \\
      & SD3  + LUA & 184.58          & 103.94          & 0.0022          & 0.0083          & 0.768 & 20.292 \\
    \midrule
    \multirow{3}{*}{$\times4$}
      & FLUX + LUA & \textbf{181.06} & \textbf{62.30}  & 0.0018          & \textbf{0.0085} & \textbf{0.772} & 31.908 \\
      & SDXL + LUA & 182.42          & 71.92           & \textbf{0.0015} & 0.0152          & 0.754 & \textbf{6.87} \\
      & SD3  + LUA & 183.34          & 67.25           & 0.0016          & 0.0095          & 0.769 & 21.843 \\
    \bottomrule
  \end{tabular*}
\end{table*}

\subsection{Qualitative Results}
\label{sec:qual_results}

\begin{figure*}[t]
  \centering
  \includegraphics[width=0.91\textwidth]{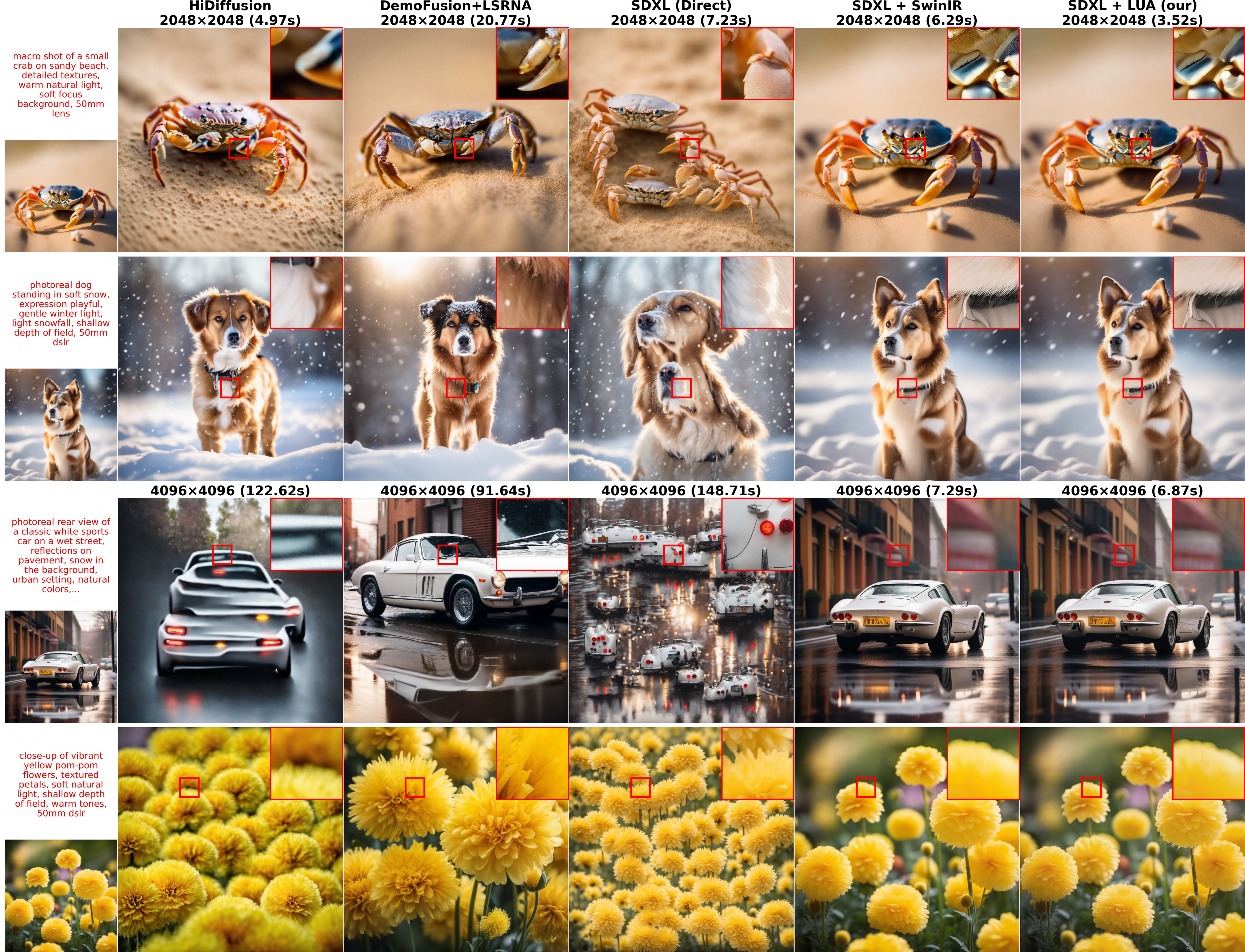}
  \caption{Qualitative comparison at $2048^2$ and $4096^2$ starting from the same $1024^2$ SDXL base generations. Each row uses identical seeds and prompts (GPT-generated captions from OpenImages validation). Red boxes indicate $12\times$ magnified crops; titles report per-image runtime. For visual clarity we show the DemoFusion+LSRNA variant in place of plain DemoFusion. The column with \emph{SDXL+LUA (ours)} achieves the lowest latency and produces clean, stable textures without the high-resolution artifacts (e.g., repetition, gridding) seen in direct high-res sampling, and without the sharpening noise typical of pixel-space SR.}
  \label{fig:qual_comp_two_col}
\end{figure*}

Figure~\ref{fig:qual_comp_two_col} presents side-by-side comparisons at $2048^2$ (top two rows) and $4096^2$ (bottom two), starting from the same $1024^2$ SDXL bases (identical seeds/prompts). Direct high-resolution sampling (\emph{SDXL Direct}) exhibits canonical large-scale failures: duplicated structures and geometry drift in the crab legs and sand granules (row~1), brittle clumps in dog fur and snow particles (row~2), and warped taillights/reflections in the street scene (row~3). \emph{HiDiffusion} shows similar breakdowns at $4$K (row~3), indicating that training-free escalation struggles to maintain global layout at extreme scales. In contrast, SR-from-base approaches avoid these hallucinations by inheriting the cleaner $1024^2$ generation.

Among SR methods, pixel-space SwinIR sharpens but introduces ringing/haloing and plastic textures: overshot specular ridges on the crab shell (row 1), halos at fur boundaries (row 2), glare around car edges (row 3), and granular noise on petals (row 4). DemoFusion+LSRNA restores rich texture but requires a second diffusion stage with much higher latency. SDXL+LUA (ours) preserves edge continuity and microstructure with fewer artifacts: crisp eyelashes and shell ridges without halos (row 1), distinct yet stable fur strands (row 2), sharp panel seams and reflections without brittleness (row 3), and coherent high-frequency petal detail (row 4). Per-image runtimes overlaid above each panel align with Table~\ref{tab:quantitative_results}: LUA attains comparable or better visual quality at the lowest latency via single-decode latent upscaling. Additional examples are provided in the appendix.

\subsection{Ablation Studies}
\label{sec:ablations}

\paragraph{Multi-stage training effectiveness.}
\label{sec:abl-multi-stage}
We assess the curriculum by removing individual stages and comparing against a latent $\ell_1$–only baseline. Table~\ref{tab:abl_multi_stage} shows that the full configuration (I+II+III) yields the best fidelity at both scales. At $\times2$, the full model attains PSNR 32.54 / LPIPS 0.138 versus 28.96 / 0.172 for I+II (w/o III). At $\times4$, II+III (w/o I) reaches 27.40 / 0.192 compared to 27.94 / 0.184 for the full model. Removing Stage~III reduces edge sharpness and perceptual quality; removing Stage~II weakens latent–pixel consistency and increases artifacts. A pixel-only variant fails to converge with a frozen VAE. Please, follow appendix for training settings details.

\begin{table}[t]
  \centering
  \scriptsize
  \setlength{\tabcolsep}{5pt}%
  \renewcommand{\arraystretch}{1.12}
  \setlength{\aboverulesep}{0.2ex}%
  \setlength{\belowrulesep}{0.2ex}%
  \setlength{\extrarowheight}{0.3ex}
  \caption{Ablation on the multi-stage latent–pixel curriculum for $\,\times2$ ($512{\rightarrow}1024$) and $\,\times4$ ($256{\rightarrow}1024$). Best results are in \textbf{bold}.}
  \vspace{-2pt}
  \label{tab:abl_multi_stage}
  \begin{tabular*}{\linewidth}{@{\extracolsep{\fill}} l cc cc}
    \toprule
    Configuration & PSNR $\uparrow$ & LPIPS $\downarrow$ & PSNR $\uparrow$ & LPIPS $\downarrow$ \\
    \midrule
    Latent $\ell_1$   & 28.53 & 0.198 & 26.16 & 0.236 \\
    I+II (w/o III)    & 28.96 & 0.172 & 26.67 & 0.213 \\
    I+III (w/o II)    & 31.05 & 0.150 & 27.10 & 0.198 \\
    II+III (w/o I)    & 31.60 & 0.145 & 27.40 & 0.192 \\
    Full (I+II+III)   & \textbf{32.54} & \textbf{0.138} & \textbf{27.94} & \textbf{0.184} \\
    \bottomrule
  \end{tabular*}
  \vspace{-4pt}
\end{table}

\paragraph{Multi-scale super-resolution adaptation.}
\label{sec:abl_multi-scale}
We compare three design choices per scale factor: (i) an implicit, coordinate-based upsampler (LIIF), (ii) per-scale specialist models trained only for a single factor (\,$\times2$ or $\times4$), and (iii) a joint multi-head model that shares a backbone and uses scale-specific heads. Table~\ref{tab:abl_multi_scale} reports within-scale results. The joint multi-head design attains the best PSNR/LPIPS at both $\times2$ and $\times4$, while also consolidating capacity in a single backbone, reducing storage and training overhead (see appendix for resource analysis and experiments settings).

\begin{table}[t]
  \centering
  \scriptsize
  \setlength{\tabcolsep}{4.2pt}%
  \renewcommand{\arraystretch}{1.02}
  \setlength{\aboverulesep}{0.2ex}%
  \setlength{\belowrulesep}{0.2ex}%
  \setlength{\extrarowheight}{0.3ex}
  \caption{Ablation of upsampling strategies for $\,\times2$ ($512{\rightarrow}1024$) and $\,\times4$ ($256{\rightarrow}1024$). Best results are in \textbf{bold}.}
  \vspace{-2pt}
  \label{tab:abl_multi_scale}
  \begin{tabular*}{\linewidth}{@{\extracolsep{\fill}} l cc cc}
    \toprule
    Variant & PSNR $\uparrow$ & LPIPS $\downarrow$ & PSNR $\uparrow$ & LPIPS $\downarrow$ \\
    \midrule
    LIIF                 & 29.10 & 0.210 & 26.10 & 0.235 \\
    Separated-$\times2$  & 31.92 & 0.150 &  --   &  --   \\
    Separated-$\times4$  &  --   &  --   & 27.71 & 0.189 \\
    Joint Multi-Head     & \textbf{32.54} & \textbf{0.138} & \textbf{27.94} & \textbf{0.184} \\
    \bottomrule
  \end{tabular*}
  \vspace{-4pt}
\end{table}

\section{Discussion}
Our approach inherits limitations from its adapter nature: errors or biases in the generator’s latent are faithfully upscaled, so artifacts in the base sample can persist at higher resolution. A promising direction is joint refinement–and–upscaling directly in latent space before decoding, e.g., lightweight consistency modules to suppress artifacts while preserving semantics, with uncertainty-aware gating to invoke refinement only when needed. Beyond text-to-image, the same mechanism can serve image-to-image tasks requiring high-resolution outputs—such as depth-to-image or semantic maps, where preserving structure during upscaling is critical. Finally, extending the adapter to video with temporal consistency (e.g., recurrent latent refinement or temporal priors) remains essential for practical high-resolution synthesis in dynamic settings.

\section{Conclusion}
We introduced LUA, a single-pass latent upscaler inserted between a pretrained generator and a frozen VAE decoder, and demonstrated that latent-space upscaling is trainable via a multi-stage latent–pixel curriculum with scale-specific $\times2/\times4$ heads. On OpenImages, at $2048^2$ and $4096^2$, SDXL+LUA achieves state-of-the-art single-decode fidelity (FID 180.80 / 176.90; pFID 97.90 / 61.80) while remaining the fastest; at $2048^2$ it runs in 3.52\,s versus 7.23\,s for SDXL~(Direct), outperforms pixel-space SR (SDXL+SwinIR), and approaches multi-stage re-diffusion quality at a fraction of the runtime. At $1024^2$, performance trails the strongest baselines due to the $64{\times}64$ input latent constraint, though patch-level fidelity remains competitive. A single backbone transfers across SDXL, SD3, and FLUX with only a first-layer channel change and brief fine-tuning, demonstrating robust cross-VAE and multi-scale generalization without modifying the generator or adding diffusion stages. Overall, these results establish single-decode latent upscaling as a practical alternative to multi-stage high-resolution pipelines.

\section*{Acknowledgements}

The authors are grateful to colleagues at CAIRO, Technical University of Applied Sciences Würzburg–Schweinfurt, Germany, for their sustained support throughout this project. We especially thank Pavel Chizhov for insightful discussions and assistance with manuscript revisions, which substantially improved the quality and clarity of this work.

{\small
\bibliographystyle{ieee_fullname} 
\bibliography{paper}               
}

\clearpage
\appendix
\clearpage
\appendix

\twocolumn[{%
\centering
\Large\bfseries
One Small Step in Latent, One Giant Leap for Pixels:\\
Fast Latent Upscale Adapter for Your Diffusion Models\\[0.6em]
\normalfont\large
Supplementary Material\par
\vspace{1em}
}]

\section{Training and Resource Analysis}
\label{sec:appendix-training}

This section details how LUA is trained in practice. We first specify the
exact loss compositions and weights used in the three-stage curriculum
(Sec.~\ref{sec:appendix-loss-weights}), then summarize the optimization and
scheduler settings (Sec.~\ref{sec:appendix-optim}), and finally discuss why
the staged design is important for stability with a frozen VAE decoder
(Sec.~\ref{sec:appendix-latent-stability}). We conclude with the
multi-scale training protocol and the associated resource usage
(Sec.~\ref{sec:appendix-multiscale-train}).

\subsection{Per-stage Loss Weights}
\label{sec:appendix-loss-weights}

\begin{figure}[t]
    \centering
    \includegraphics[width=\linewidth]{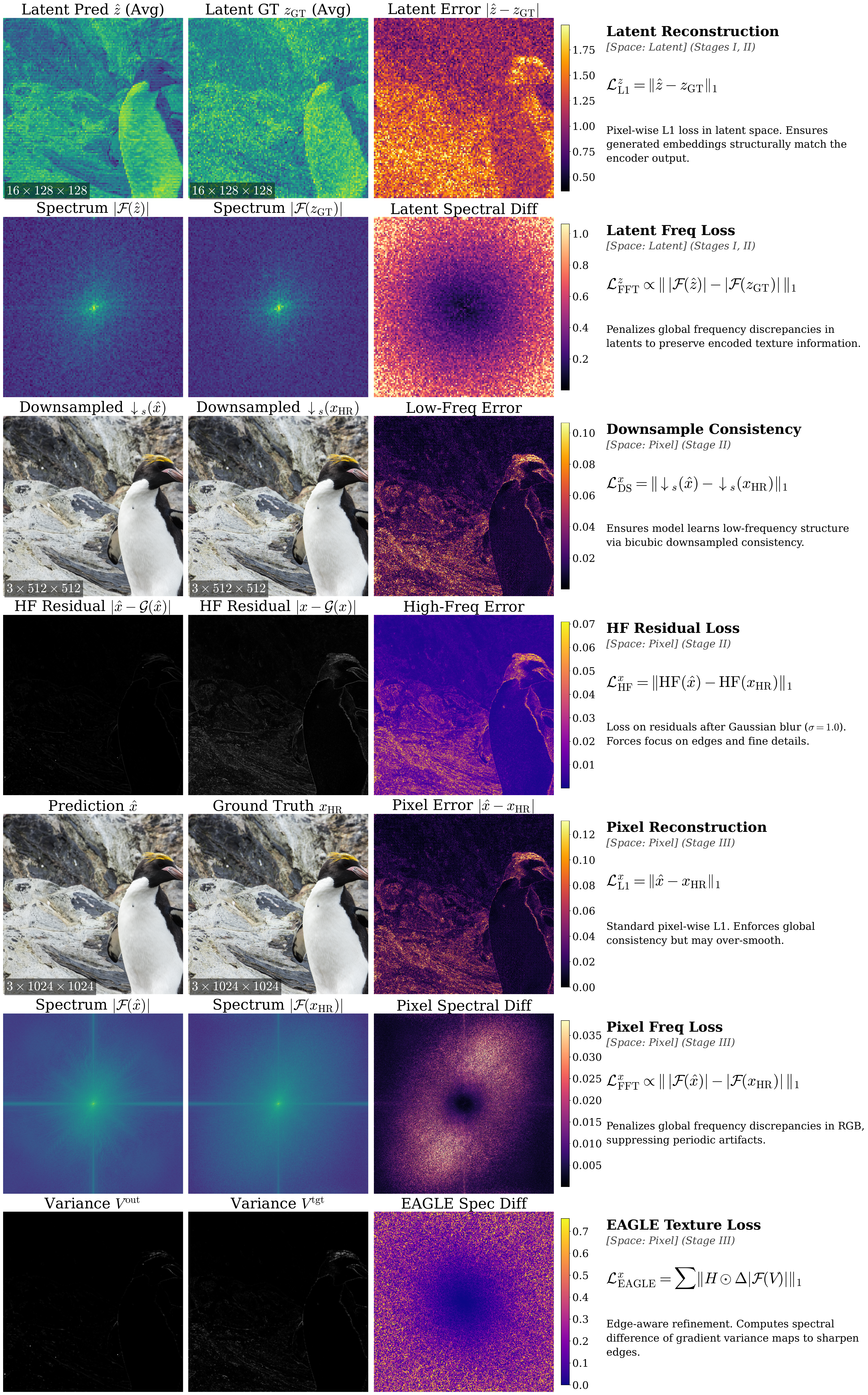}
    \caption{Visualization of all losses in the three-stage curriculum:
    latent reconstruction/frequency (Rows~1--2), downsample consistency and
    high-frequency residuals (Rows~3--4), and pixel reconstruction,
    pixel frequency, and EAGLE texture (Rows~5--7). Columns show prediction,
    ground truth, the corresponding error map, and the loss formula; brighter
    regions indicate larger gradient contributions.}
    \label{fig:losses_visualization}
\end{figure}

Let $z$ and $\hat{z}$ denote the input and upscaled latents, and
$z_{\mathrm{HR}}$ the reference high-resolution latent (from the frozen
FLUX encoder). Let $x = D(z)$, $\hat{x} = D(\hat{z})$, and
$x_{\mathrm{HR}} = D(z_{\mathrm{HR}})$ be the corresponding decoded
images via the frozen decoder $D$. Superscripts ${}^{z}$ and ${}^{x}$
indicate whether a loss is computed in latent or pixel space,
respectively. For each scale factor $\alpha \in \{2,4\}$ we use identical
coefficients. Figure~\ref{fig:losses_visualization} provides a visual
overview of all objective terms, their error maps, and their usage across
stages.

\paragraph{Stage I (latent-domain alignment).}
Stage~I uses the latent $\ell_1$ and latent FFT losses from
Eqs.~(8)–(9) for each scale:
\begin{equation}
\mathcal{L}_{\mathrm{SI}}^{(\alpha)}
= \alpha_1 \, \mathcal{L}_{\mathrm{L1}}^{z,(\alpha)}
+ \beta_1  \, \mathcal{L}_{\mathrm{FFT}}^{z,(\alpha)},
\end{equation}
where $\mathcal{L}_{\mathrm{L1}}^{z}$ is the element-wise $\ell_1$ distance
between $\hat{z}$ and $z_{\mathrm{HR}}$, and
$\mathcal{L}_{\mathrm{FFT}}^{z}$ is the channel-wise FFT magnitude loss in
latent space. We set
\[
\alpha_1 = 1.0, \qquad
\beta_1  = 0.1,
\]
so that latent reconstruction is the primary term and spectral alignment
acts as a lower-weight regularizer. Rows~1--2 of
Fig.~\ref{fig:losses_visualization} visualize these losses in latent space.

\paragraph{Stage II (joint latent--pixel consistency).}
Stage~II augments the Stage~I objective with pixel-domain terms
(Eqs.~(11)–(12)):
\begin{equation}
\mathcal{L}_{\mathrm{SII}}^{(\alpha)}
= \alpha_2 \, \mathcal{L}_{\mathrm{L1}}^{z,(\alpha)}
+ \beta_2  \, \mathcal{L}_{\mathrm{FFT}}^{z,(\alpha)}
+ \gamma_2 \, \mathcal{L}_{\mathrm{DS}}^{x,(\alpha)}
+ \delta_2 \, \mathcal{L}_{\mathrm{HF}}^{x,(\alpha)}.
\end{equation}
Here, $\mathcal{L}_{\mathrm{DS}}^{x}$ is the bicubic downsample-consistency
loss between $\hat{x}$ and $x_{\mathrm{HR}}$, computed after downsampling
both images by a factor of $2$; and $\mathcal{L}_{\mathrm{HF}}^{x}$ is the
Gaussian high-frequency $\ell_1$ loss (kernel size $5$, $\sigma = 1.0$) on
the residuals $(x - \mathcal{G}_\sigma(x))$. We use
\[
\alpha_2 = 1.0, \quad
\beta_2  = 0.1, \quad
\gamma_2 = 0.1, \quad
\delta_2 = 0.05.
\]
Rows~3--4 of Fig.~\ref{fig:losses_visualization} show the corresponding
downsample-consistency and high-frequency residual error maps.

\paragraph{Stage III (edge-aware pixel refinement).}
Stage~III operates only in pixel space as in Eqs.~(14)–(16):
\begin{equation}
\mathcal{L}_{\mathrm{SIII}}^{(\alpha)}
= \alpha_3 \, \mathcal{L}_{\mathrm{L1}}^{x,(\alpha)}
+ \beta_3  \, \mathcal{L}_{\mathrm{FFT}}^{x,(\alpha)}
+ \gamma_3 \, \mathcal{L}_{\mathrm{EAGLE}}^{x,(\alpha)},
\end{equation}
where $\mathcal{L}_{\mathrm{L1}}^{x}$ is the pixel-wise $\ell_1$ loss
between $\hat{x}$ and $x_{\mathrm{HR}}$, and
$\mathcal{L}_{\mathrm{FFT}}^{x}$ is the FFT magnitude loss in image space
(defined analogously to $\mathcal{L}_{\mathrm{FFT}}^{z}$ but applied to
RGB images). The edge-aware term $\mathcal{L}_{\mathrm{EAGLE}}^{x}$ follows
Sun \emph{et al.}~\cite{sun2025eagle} and is defined as
\begin{equation}
\begin{aligned}
\mathcal{L}_{\mathrm{EAGLE}}^{x}
&= \big\lVert H \odot \big(
    |\mathcal{F}(V_x^{\mathrm{out}})|
    - |\mathcal{F}(V_x^{\mathrm{tgt}})|
  \big) \big\rVert_1 \\
&\quad
 + \big\lVert H \odot \big(
    |\mathcal{F}(V_y^{\mathrm{out}})|
    - |\mathcal{F}(V_y^{\mathrm{tgt}})|
  \big) \big\rVert_1,
\end{aligned}
\end{equation}
where $\mathcal{F}(\cdot)$ is the 2D FFT,
$V_x^{\mathrm{out}}$ and $V_x^{\mathrm{tgt}}$ are the per-patch variance
maps of the horizontal image gradients of $\hat{x}$ and $x_{\mathrm{HR}}$
(and analogously for $V_y^{\mathrm{out}}, V_y^{\mathrm{tgt}}$ in the
vertical direction), and $H$ is a Gaussian high-pass mask in the frequency
domain,
\begin{equation}
H(u,v) = 1 - \exp\!\left(
    -\frac{u^2 + v^2}{2 f_c^2}
\right),
\end{equation}
with normalized spatial frequencies $(u,v)$ and cutoff $f_c = 0.5$.
Gradients are computed with $3{\times}3$ Scharr filters, and $V_x, V_y$
are obtained by non-overlapping variance pooling over $3{\times}3$
patches. In our implementation, EAGLE is applied to luminance for RGB
images and reduced with an $\ell_1$ norm over all frequencies.
We use
\[
\alpha_3 = 10.0, \quad
\beta_3  = 1.0, \quad
\gamma_3 = 5\times 10^{-5},
\]
so that the strong $\ell_1$ term stabilizes Stage~III, while
$\mathcal{L}_{\mathrm{EAGLE}}^{x}$ acts as a light regularizer that
sharpens edges and suppresses grid artifacts. Rows~5--7 of
Fig.~\ref{fig:losses_visualization} illustrate the corresponding pixel
reconstruction, pixel-frequency, and EAGLE texture losses.

\paragraph{Hyperparameter selection.}
We select all coefficients
$(\alpha_k,\beta_k,\gamma_k,\delta_k)$ using a small held-out latent
dataset (10k training pairs, 100 validation images). We first perform a
coarse grid search over discrete weights
$\{0, 0.05, 0.1, 0.5, 1, 5, 10\}$, measuring a scalar objective combining
patch FID and LPIPS on the validation set. We then refine the best region
with Optuna-based TPE search, using continuous ranges
(e.g., $\alpha_2 \in [0.5,2]$, $\beta_2,\gamma_2 \in [0.05,0.5]$,
$\delta_2 \in [0.01,0.1]$). The configuration above minimizes the
validation objective averaged over both scale heads and is reused for the
full training runs.

\subsection{Optimization and Scheduler Settings}
\label{sec:appendix-optim}

All stages employ Adam with learning rate $2\times10^{-4}$ and zero weight
decay. Stage~I and Stage~II use momentum parameters
$(\beta_1,\beta_2) = (0.9, 0.995)$, while Stage~III uses
$(\beta_1,\beta_2) = (0.9, 0.99)$ to adapt more quickly once supervision
moves entirely to pixel space. Each stage is trained for $125{,}000$ steps
with a MultiStepLR scheduler (milestones at $62{,}500$, $93{,}750$, and
$112{,}500$ steps; decay factor $\gamma = 0.5$) and no warm-up. We maintain
an exponential moving average of the generator parameters with decay
$0.999$ and use the EMA weights for validation and final reporting.

To prevent gradient explosion, we apply global $\ell_2$-norm clipping with
$\text{max\_norm} = 0.4$ to the generator in all stages. Stage~I uses a
per-GPU batch size of $64$, combined with gradient accumulation to reach
an effective batch of approximately $2{,}048$ latent samples. Stages~II
and~III use a per-GPU batch size of $8$ with accumulation, leading to an
effective batch of $32$ decoded samples for the more expensive
pixel-domain losses. All experiments use distributed data parallelism
with NCCL and mixed-precision training on 8 GPUs.

\subsection{Latent Normalization and Limitations}
\label{sec:appendix-latent-stability}

Pixel-only training of LUA with a frozen VAE decoder is numerically
unstable: when supervision is applied directly in image space from
scratch, latent activations drift outside the decoder’s typical range,
leading to saturated codes, localized ``broken pixels'' after decoding,
and eventual loss divergence. The three-stage curriculum described in
Sec.~\ref{Multi-stage-train} mitigates this by first aligning latent
statistics to those of the encoder and only then transitioning to
pixel-domain optimization. In ablations, removing the intermediate
(latent–pixel) stage or training the final pixel-only stage from random
initialization consistently causes instability and pronounced grid-like
artifacts.

\begin{figure}[t]
    \centering
    \includegraphics[width=\linewidth]{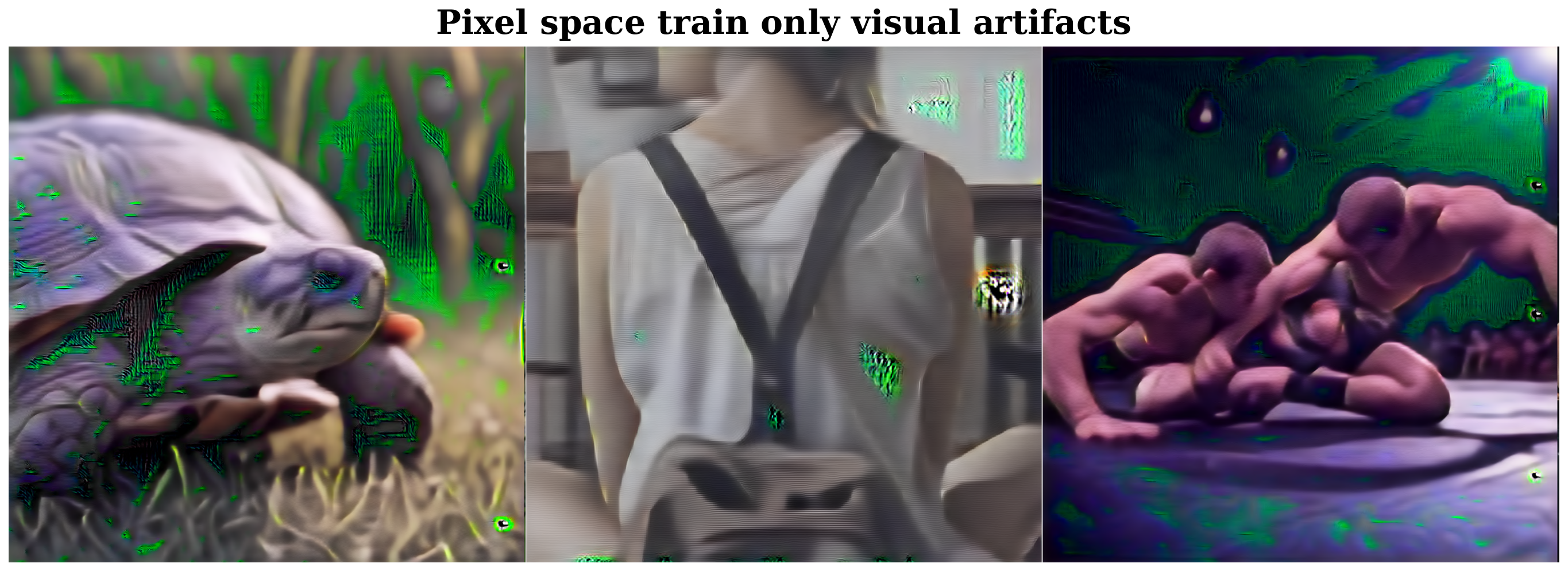}
    \caption{Artifacts from a pixel-only baseline trained without latent
    normalization. Decoded crops show saturated streaks and ``broken''
    pixels caused by out-of-range latents passed to the frozen VAE
    decoder. These failures are avoided by the proposed three-stage
    latent–pixel curriculum.}
    \label{fig:latent_break_artifacts}
\end{figure}

A remaining limitation is the absence of an explicit bound on the latent
values produced by LUA. Under out-of-distribution inputs, the adapter can
still emit latents that decode to localized corruptions
(Fig.~\ref{fig:latent_break_artifacts}). Introducing explicit
latent-domain regularization or lightweight post-processing
(e.g., bounded residual corrections) is therefore a natural direction for
improving robustness under distribution shift.

\subsection{Multi-scale Training and Resource Usage}
\label{sec:appendix-multiscale-train}

LUA is trained as a single multi-scale model that supports both
$\times2$ and $\times4$ latent upscaling. For each training sample we
form two latent pairs sharing the same high-resolution target: an
intermediate-to-high pair for the $\times2$ head and a coarse-to-high
pair for the $\times4$ head. In the early phase of each stage, both
heads are supervised at every iteration so that the shared Swin backbone
learns features useful for both scales. In the final $50{,}000$ steps,
we switch to stochastic head selection, sampling either the $\times2$ or
$\times4$ head with equal probability per iteration to encourage
head-specific specialization while keeping the backbone shared. This
protocol is reused across all three stages; only the loss composition
changes (latent-only, joint latent–pixel, and pixel-only refinement).

All experiments were run on 8 NVIDIA H100 80GB GPUs. The three stages
required approximately $16$ hours (Stage~I), $7.2$ hours (Stage~II), and
$11$ hours (Stage~III), for a total of about $34.1$ hours, i.e.,
\[
34.1 \,\text{h} \times 8 \,\text{GPUs} \approx 2.7 \times 10^{2}
\;\text{GPU-hours}.
\]
This yields a single model that handles both scale factors. Training two
independent models, one specialized for $\times2$ and one for $\times4$,
with the same three-stage schedule would roughly double the budget to
$\approx 5.4 \times 10^{2}$ GPU-hours. The multi-head design therefore
saves on the order of $2.7 \times 10^{2}$ GPU-hours (about a factor of
two in training cost) while producing a single checkpoint deployable
across both scales.

\section{Validation Prompts and Captions}
\label{sec:appendix-prompts}

This section describes how we obtain text prompts for the 1k-image
validation set and how they are used for multi-resolution evaluation.

\subsection{Prompt Generation Procedure}
\label{sec:appendix-prompt-proc}

For each image in the validation set (Sec.~\ref{sec:eval_protocol}) we
generate a single caption with a vision–language model based on GPT.
The model is queried once per image with a fixed decoding configuration
and instructed to return a short, self-contained description suitable
for text-to-image generation. We query a GPT-based captioner with a fixed, 
structured instruction:

\begin{quote}
\small
\texttt{You are an image captioning model for text-to-image diffusion.}\\
\texttt{Describe the image in 1-2 sentences, focusing on concrete objects,}\\
\texttt{their attributes, layout, and photographic style. Return JSON as}\\
\texttt{\{ "caption": "<caption here>" \}.}
\end{quote}

The resulting \texttt{"caption"} field is used as the prompt for all
methods and resolutions ($1024^2$, $2048^2$, $4096^2$). 

\subsection{Prompt Set Statistics and Examples}
\label{sec:appendix-prompt-stats}

The prompts typically consist of one or two sentences (about 25–40
tokens), mentioning the main objects, their spatial arrangement, and
basic style cues (e.g., lighting, depth of field). They cover a broad
range of content, including natural scenes, animals, indoor objects, and
people, and are used uniformly across all compared methods.

\begin{figure*}[t]
  \centering
  \includegraphics[width=1.0\textwidth]{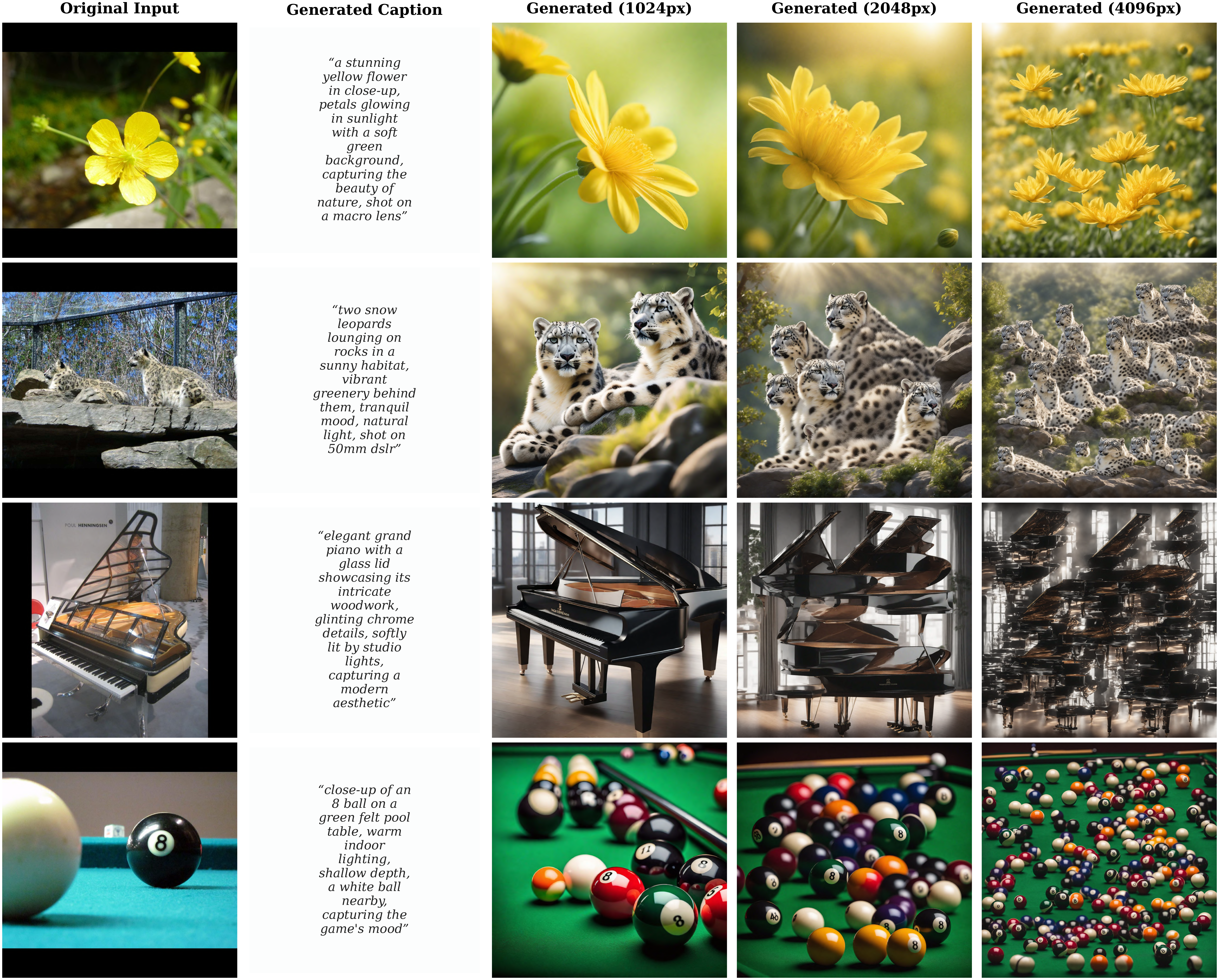}
  \caption{Caption-conditioned multi-resolution generation on the
  validation set. For each held-out photograph (left), we obtain a
  structured caption via the procedure in
  Sec.~\ref{sec:appendix-prompt-proc} (second column) and synthesize SDXL
  images at $1024^2$, $2048^2$, and $4096^2$ from the same prompt
  (right). While global semantics are preserved, native high-resolution
  sampling introduces hallucinations such as object multiplication and
  exaggerated detail, motivating the $1$K+upscaling strategy in
  Sec.~\ref{sec:appendix-1k-vs-2k4k}.}
  \label{fig:multires_qualitative}
\end{figure*}

Figure~\ref{fig:multires_qualitative} shows representative examples:
for each validation image we display the original photograph, the
generated caption, and synthesized outputs at
$1024{\times}1024$, $2048{\times}2048$, and $4096{\times}4096$
resolutions driven by the same prompt. As resolution increases, some
pipelines exhibit typical high-resolution hallucinations such as object
duplication and overly dense textures, despite the caption remaining
fixed. This observation motivates the strategy of generating a clean
base image at a moderate resolution and then applying upsampling in
latent or pixel space, rather than sampling directly at $2$K–$4$K.
We analyze this trade-off and compare direct high-resolution sampling
with 1K+LUA upscaling in Sec.~\ref{sec:appendix-1k-vs-2k4k}.

\section{Baseline and Reproduction Details}
\label{sec:appendix-baselines}

This section summarizes the main configuration choices for all baselines in
Sec.~\ref{sec:experiments} and the common evaluation protocol used to
ensure a fair comparison.

\subsection{Diffusion and Super-resolution Baselines}
\label{sec:appendix-diffusion-baselines}

Unless otherwise stated, all SDXL-based diffusion experiments use the same
caption-derived positive prompts (Sec.~\ref{sec:appendix-prompts}), a
fixed negative prompt
\begin{quote}
\small
\emph{blurry, ugly, duplicate, poorly drawn face, deformed, mosaic, artifacts, bad limbs}
\end{quote}
classifier-free guidance scale $7.5$, global seed $42$, and the
\emph{stabilityai/stable-diffusion-xl-base-1.0} checkpoint with a DDIM
sampler in half precision.

\paragraph{SDXL (Direct).}
SDXL (Direct) uses the standard SDXL pipeline without any explicit
upscaling or refinement. For each target resolution
($1024^2$, $2048^2$, $4096^2$) we sample with 30 DDIM steps and
$\eta = 1.0$.

\paragraph{HiDiffusion.}
HiDiffusion is evaluated by applying the official HiDiffusion modification
to the same SDXL pipeline. For each prompt and target resolution we use
30 DDIM steps, $\eta = 1.0$, guidance scale $7.5$, and seed $42$.

\paragraph{DemoFusion.}
DemoFusion is instantiated as a DemoFusion–SDXL pipeline using the SDXL
base model and the VAE
\emph{madebyollin/sdxl-vae-fp16-fix}. For all resolutions we use
40 sampling steps, guidance scale $7.5$, and seed $42$. DemoFusion-specific
hyperparameters are: view batch size $4$, stride $64$ pixels, cosine
scales $(3.0, 1.0, 1.0)$, and noise level $\sigma = 0.8$.

\paragraph{LSRNA--DemoFusion.}
For LSRNA--DemoFusion we augment the DemoFusion–SDXL pipeline with the
official LSRNA latent super-resolution module
(\emph{swinir-liif-latent-sdxl.pth}), keeping the SDXL backbone and DDIM
sampler unchanged and enabling tiled VAE decoding. For all resolutions we
use 50 DDIM steps, guidance scale $7.5$, and seed $42$. Additional
hyperparameters are: view batch size $8$, stride ratio $0.5$, cosine
scales $(3.0, 1.0, 1.0)$, noise level $\sigma = 0.8$, latent standard
deviation range $[0.0, 1.2]$, and inversion depth $30$.

\paragraph{FLUX backbone.}
For the FLUX experiments we use \emph{FLUX.1-Krea-dev} as implemented in
the public FluxPipeline. For each prompt and target resolution, FLUX
latents are generated with 12 denoising steps, guidance scale $4.5$, and
bfloat16 precision, and decoded with the native FLUX VAE. These latents
serve as inputs to LUA for the FLUX-based evaluations in
Sec.~\ref{sec:experiments}.

\paragraph{Pixel-space super-resolution (SwinIR).}
For pixel-space super-resolution we start from the official SwinIR
checkpoints for $\times2$ and $\times4$ SR and fine-tune them on the same
high-resolution image collection as LUA using the official training
hyperparameters, with bicubically downsampled inputs and a combination of
$\ell_1$ and perceptual losses. At test time, SDXL first generates a
$1024^2$ image, which is then either upscaled to $2048^2$ by the
$\times2$ SwinIR model or to $4096^2$ by the $\times4$ SwinIR model,
depending on the target resolution. No additional post-processing is
applied.

\subsection{Shared Evaluation Protocol}
\label{sec:appendix-fairness}

To make the comparison as controlled as possible, all methods share the
following settings:

\begin{itemize}
    \item \textbf{Prompts.} For each validation image, all methods at all
    resolutions use the same caption-derived prompt from
    Sec.~\ref{sec:appendix-prompts} together with the fixed negative
    prompt above.
    \item \textbf{Backbone and guidance.} All SDXL-based diffusion methods
    use the same SDXL checkpoint, DDIM sampler, guidance scale $7.5$, and
    mixed-precision computation; the only differences lie in the
    high-resolution strategy (direct sampling, HiDiffusion, DemoFusion,
    LSRNA, or LUA).
    \item \textbf{Randomness.} For SDXL-based methods, we fix the seed to
    $42$ for each method and resolution, ensuring deterministic sampling
    and identical noise realizations across baselines.
    \item \textbf{Runtime measurement.} For diffusion and SR pipelines we
    measure the wall-clock time of a full forward pass per image and
    report the mean per-image time for each resolution, as in
    Sec.~\ref{sec:experiments}.
\end{itemize}

\section{Additional Ablations}
\label{sec:appendix-ablations}

This section provides an additional analysis that complements the ablation
studies in Sec.~\ref{sec:ablations}, focusing on the statistical
differences between latent and pixel super-resolution and how LUA behaves
with respect to these distributions.

\subsection{Latent vs.\ Pixel Super-resolution Statistics}
\label{sec:appendix-latent-vs-pixel}

\begin{figure}[t]
    \centering
    \includegraphics[width=\linewidth]{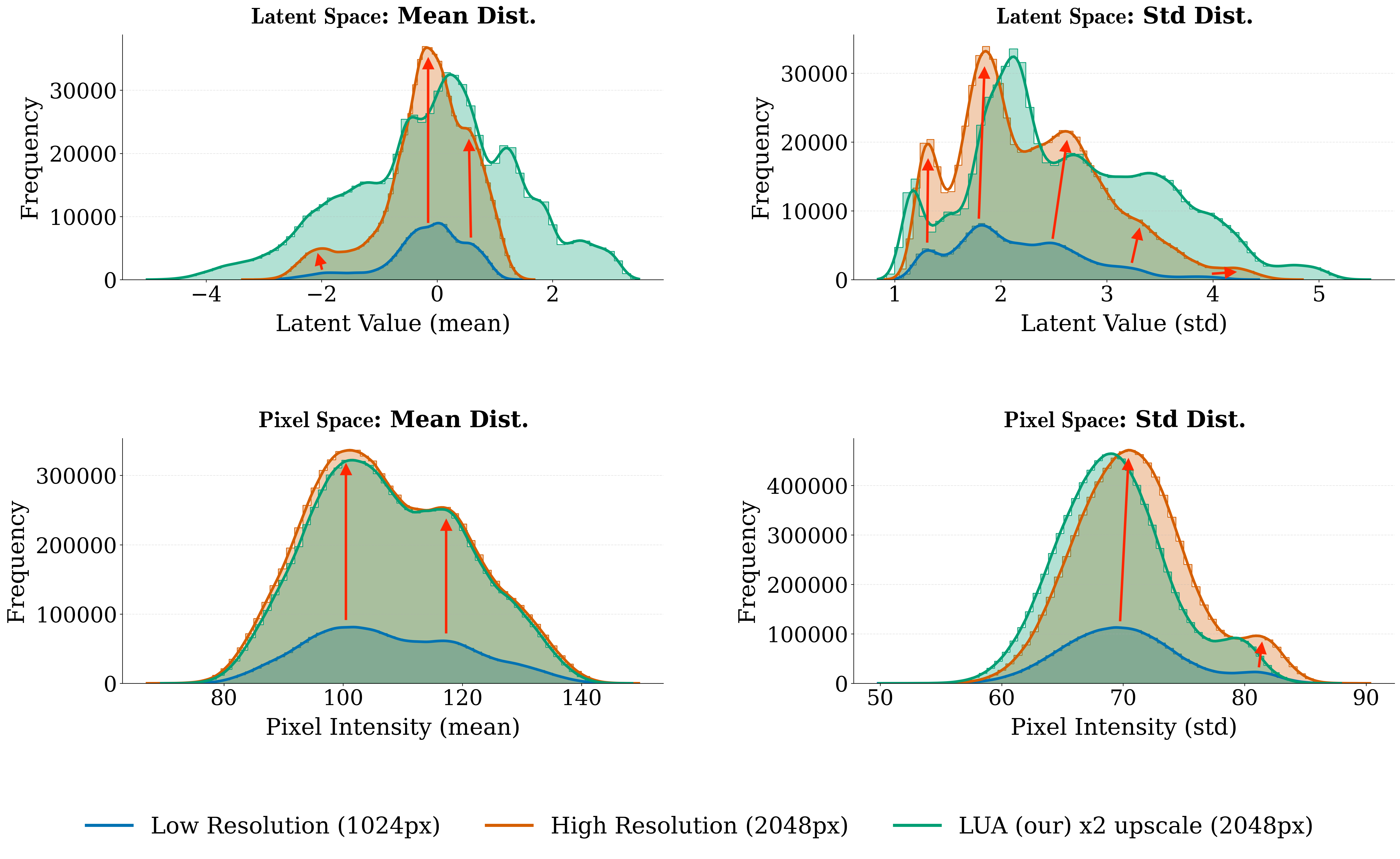}
    \caption{KDEs of channel-aggregated mean and standard deviation in
    latent (top) and pixel (bottom) domains for low-resolution inputs
    (blue), high-resolution targets (orange), and LUA outputs (green).
    Latent distributions are complex and multimodal, with low-resolution
    latents collapsing high-resolution modes (red arrows), whereas pixel 
    distributions remain simple and stable.}
    \label{fig:distributions}
\end{figure}

We analyze whether latent super-resolution is statistically more
demanding than pixel-space super-resolution and how far LUA shifts
low-resolution statistics toward high-resolution ones. For each
validation image we compute channel-aggregated means and standard
deviations in latent space (per latent channel) and in pixel space (per
RGB channel). These per-image statistics are pooled over the validation
set, and kernel density estimation (KDE) is used to estimate probability
densities for low-resolution (LR) inputs, high-resolution (HR) targets,
and LUA outputs. Figure~\ref{fig:distributions} shows the resulting
distributions; Table~\ref{tab:tab_supplementary_abl_distributions}
quantifies the LR–HR gap using Wasserstein distance ($W_1$),
Jensen--Shannon divergence (JSD), and relative variance gap
($\Delta\sigma^2$), capturing global shift, density-shape differences,
and second-order mismatch, respectively.

In latent space (top row of Fig.~\ref{fig:distributions}), HR latents
exhibit rich, multimodal structure with several distinct peaks and
shoulders, while LR latents are much flatter, collapsing many modes into
broad envelopes and indicating a marked loss of structural specificity at
lower resolution. In pixel space (bottom row), LR and HR distributions
retain almost identical shapes and differ mainly by small shifts in mean
and scale, suggesting that pixel SR operates in a comparatively smooth
statistical regime.

Table~\ref{tab:tab_supplementary_abl_distributions} makes this contrast
explicit: the Wasserstein distance between LR and HR distributions is
about $40\times$ larger in latent space ($0.119$) than in pixel space
($0.003$), with JSD and variance gap following the same trend. Thus,
latent SR must reconstruct missing modes of the latent manifold, whereas
pixel SR primarily refines an already well-aligned distribution.

LUA’s distributions (green curves in Fig.~\ref{fig:distributions}) do not
perfectly match the HR ones, which is expected because LR inputs are
obtained by downscaling HR generations to enforce identical semantics,
rather than arising from a natural generation process. Within this setting,
LUA nonetheless shifts LR latents closer to the HR manifold, and in pixel
space the remaining discrepancies are further reduced by the VAE decoder,
which smooths residual latent errors. Overall, latent SR involves a much
larger distributional shift than pixel SR; LUA partially narrows this gap
and leverages the VAE to absorb remaining inconsistencies in image space.

\begin{table}[t]
    \centering
    \scriptsize
    \setlength{\tabcolsep}{4.2pt}%
    \renewcommand{\arraystretch}{1.02}%
    \setlength{\aboverulesep}{0pt}%
    \setlength{\belowrulesep}{0pt}%
    \setlength{\extrarowheight}{0pt}%
    \caption{LR--HR domain gap in pixel and latent spaces. We report
    Wasserstein distance $W_1$, Jensen--Shannon divergence (JSD), and
    relative variance gap $\Delta\sigma^2$ between low- and
    high-resolution statistics (Sec.~\ref{sec:appendix-latent-vs-pixel}).
    Latent space exhibits a substantially larger shift across all
    metrics.}
    \label{tab:tab_supplementary_abl_distributions}
    \begin{tabular*}{\linewidth}{@{\extracolsep{\fill}} lccc}
        \toprule
        Domain        & $W_1 \downarrow$ & JSD $\downarrow$ & Variance gap \\
        \midrule
        Pixel space   & 0.003            & 0.019            & 1.1\%        \\
        Latent space  & \textbf{0.119}   & \textbf{0.044}   & \textbf{6.4\%} \\
        \bottomrule
    \end{tabular*}
\end{table}

\subsection{Hallucinations in High-resolution Sampling}
\label{sec:appendix-1k-vs-2k4k}

Diffusion models are known to exhibit visual hallucinations when sampled
at very high resolutions, including duplicated structures, malformed
anatomy, and physically implausible geometry. We quantify how such
artifacts scale with resolution by auditing SDXL as a representative
text-to-image backbone.

We generate 100 images at each target resolution
$1024^2$, $2048^2$, and $4096^2$ using the caption-derived prompts from
Sec.~\ref{sec:appendix-prompts}. Each image is analyzed by a
GPT-based vision–language auditor in a two-step zero-shot protocol:
(i) determine whether any non-trivial hallucination is present; (ii) if
so, assign one or more labels from a fixed taxonomy. The auditor operates 
with the following fine-grained artifact types:

\begin{figure}[t]
    \centering
    \includegraphics[width=\linewidth]{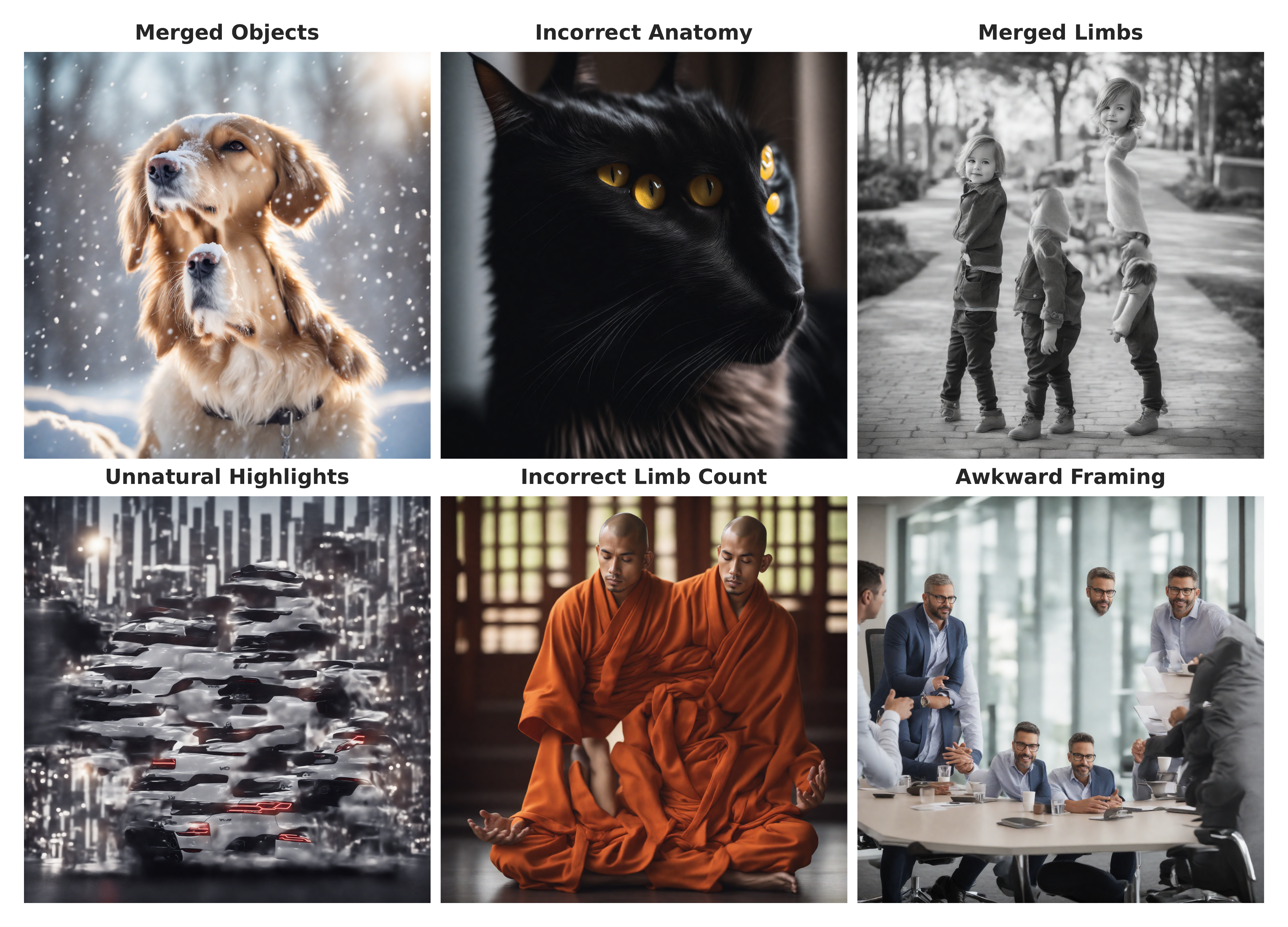}
    \caption{Representative hallucinations in native high-resolution SDXL
    sampling. For SDXL generations at $2048^2$ and $4096^2$, we show cropped
    examples corresponding to the most frequent artifact types identified by
    the auditor: Merged Objects, Incorrect Anatomy, Merged Limbs, Unnatural
    Highlights, Incorrect Limb Count, and Awkward Framing. These crops
    illustrate characteristic failure modes that become increasingly common
    as the native sampling resolution grows
    (Table~\ref{tab:tab_supplementary_abl_hallucinations}).}
    \label{fig:hallucination_examples}
\end{figure}

{\setlist[description]{nosep,leftmargin=1.1em}
\begin{description}
    \item[Incorrect Anatomy:] Unrealistic body shape or pose.
    \item[Merged Limbs:] Limbs fused or intersecting unnaturally.
    \item[Merged Objects:] Separate objects blending into one.
    \item[Distorted Face:] Strongly warped facial structure.
    \item[Blurry Background:] Background blur beyond intent.
    \item[Incorrect Limb Count:] Too many or too few limbs.
    \item[Awkward Framing:] Unnatural crop or viewpoint.
    \item[Malformed Hands:] Implausible hand or finger shapes.
    \item[Inconsistent Shadows:] Shadows contradict lighting.
    \item[Unnatural Highlights:] Physically implausible highlights.
    \item[Minor Inconsistencies:] Small local visual errors.
    \item[Impossible Perspective:] Geometrically impossible view.
    \item[Incoherent BG:] Background inconsistent with the scene.
    \item[Cluttered:] Overly dense layout hiding the subject.
    \item[Floating Objects:] Objects not supported by anything.
    \item[Blurry Patches:] Local unexpected blur in sharp areas.
    \item[Over-smooth:] Region overly smooth, missing texture.
\end{description}
}

In this audit, no hallucinations are detected at $1024^2$, whereas
9\,\% of images at $2048^2$ and 19\,\% at $4096^2$ contain at least one
artifact. Table~\ref{tab:tab_supplementary_abl_hallucinations} reports,
for each fine-grained type, how many images at each resolution receive
that label; the most frequent failures are Merged Objects, Incorrect
Anatomy, and Merged Limbs, followed by Unnatural Highlights and Incorrect
Limb Count, with representative examples in
Fig.~\ref{fig:hallucination_examples}. This pattern is consistent with
common training regimes in which diffusion backbones are optimized
primarily at moderate resolutions and only weakly exposed to very large
image sizes, so native sampling at $2$K–$4$K occurs outside the best
calibrated regime and exhibits a higher rate and diversity of
hallucinations. These findings support a two-stage design in which images
are first generated near $1024^2$ and then upscaled by a dedicated
super-resolution module; our latent upscaler adapter follows this
strategy, leveraging stable low-resolution generation and performing
resolution enhancement in latent space to reduce high-resolution
hallucinations relative to direct sampling.

\begin{table}[t]
  \centering
  \scriptsize
  \setlength{\tabcolsep}{4.0pt}%
  \renewcommand{\arraystretch}{1.05}%
  \setlength{\aboverulesep}{0pt}%
  \setlength{\belowrulesep}{0pt}%
  \setlength{\extrarowheight}{0pt}%
  \caption{Fine-grained hallucination counts vs.\ sampling resolution.
  For each artifact type, we report the number of images (out of 100) at
  each resolution that are flagged by the auditor; multiple labels may
  be assigned to the same image. The \textit{Overall} row aggregates the
  total number of hallucination tags, illustrating how both the frequency
  and variety of artifacts increase with resolution.}
  \label{tab:tab_supplementary_abl_hallucinations}
  \begin{tabular*}{\linewidth}{@{\extracolsep{\fill}} lcccc}
    \toprule
    Artifact type            & 1024px & 2048px & 4096px & Total \\
    \midrule
    Merged Objects           & 0 &  8 &  8 & 16 \\
    Incorrect Anatomy        & 0 &  6 &  7 & 13 \\
    Merged Limbs             & 0 &  5 &  6 & 11 \\
    Unnatural Highlights     & 0 &  0 &  8 &  8 \\
    Incorrect Limb Count     & 0 &  3 &  4 &  7 \\
    Awkward Framing          & 0 &  2 &  3 &  5 \\
    Inconsistent Shadows     & 0 &  1 &  2 &  3 \\
    Distorted Face           & 0 &  1 &  1 &  2 \\
    Malformed Hands          & 0 &  1 &  1 &  2 \\
    Minor Inconsistencies    & 0 &  0 &  2 &  2 \\
    Impossible Perspective   & 0 &  0 &  2 &  2 \\
    Incoherent Background    & 0 &  0 &  2 &  2 \\
    Blurry Background        & 0 &  1 &  0 &  1 \\
    Cluttered                & 0 &  0 &  1 &  1 \\
    Floating Objects         & 0 &  0 &  1 &  1 \\
    Blurry Patches           & 0 &  0 &  1 &  1 \\
    Unnatural Smoothness     & 0 &  0 &  1 &  1 \\
    \midrule
    Overall                  & 0 & 28 & 50 & 78 \\
    \bottomrule
  \end{tabular*}
\end{table}

\section{Additional FLUX-based Visualizations}
\label{sec:appendix-flux-visuals}

\begin{figure*}[t]
    \centering
    \includegraphics[width=\textwidth]{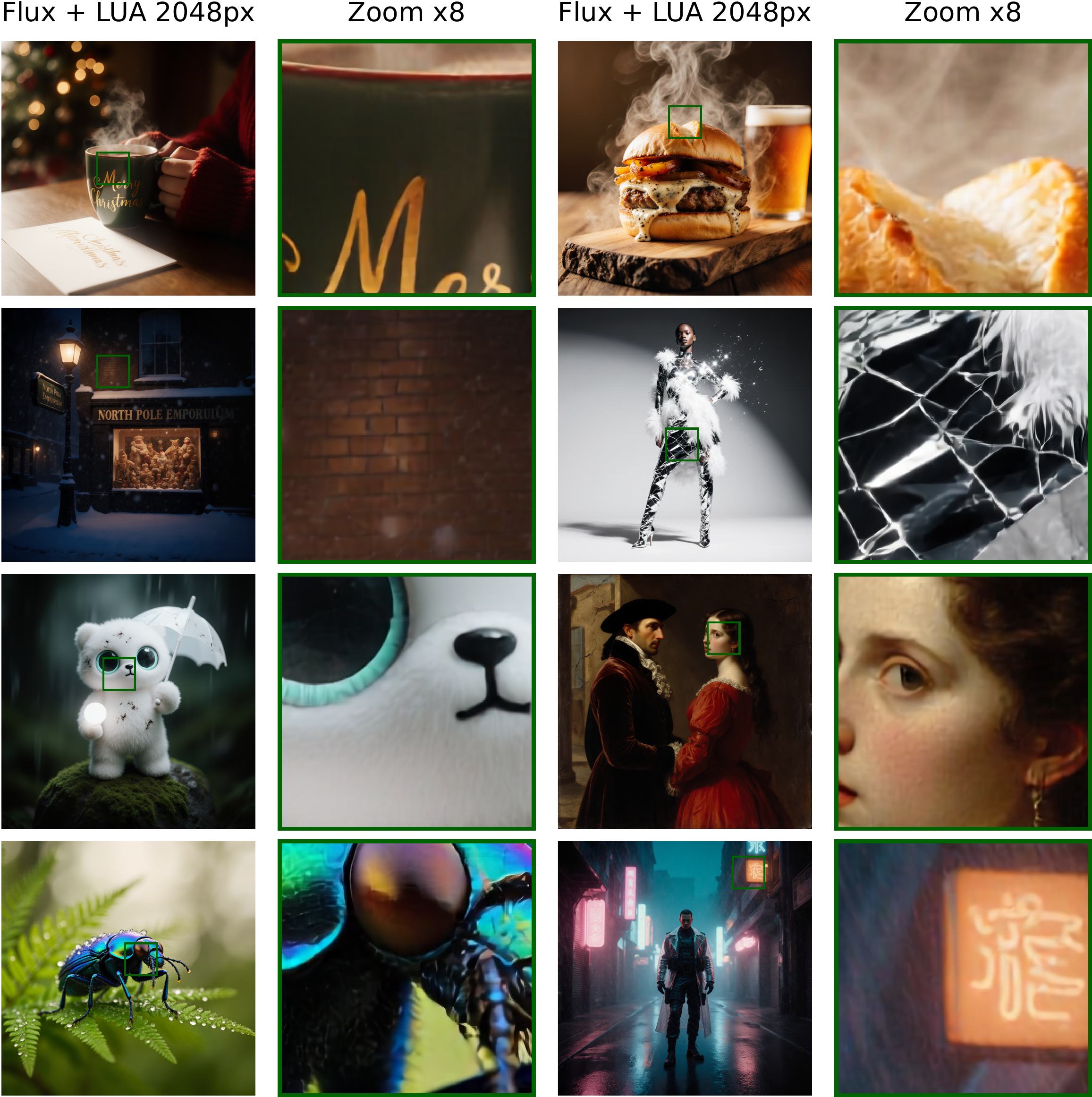}
    \caption{FLUX+LUA qualitative examples at $2048^2$ (set~1). For each
    prompt, FLUX generates a $128{\times}128$ latent (12 denoising steps),
    LUA upsamples it to $256{\times}256$, and the native FLUX VAE decodes
    once to a $2048{\times}2048$ image. Insets show $\times8$ zooms
    highlighting fine structures (hair, fabric, foliage) preserved by
    latent upscaling. Timings (generation: 46.57\,s; upscaling: 0.63\,s;
    decoding: 3.40\,s) are measured on a single NVIDIA L40S GPU.}
    \label{fig:flux_lua_examples_1}
\end{figure*}

\begin{figure*}[t]
    \centering
    \includegraphics[width=\textwidth]{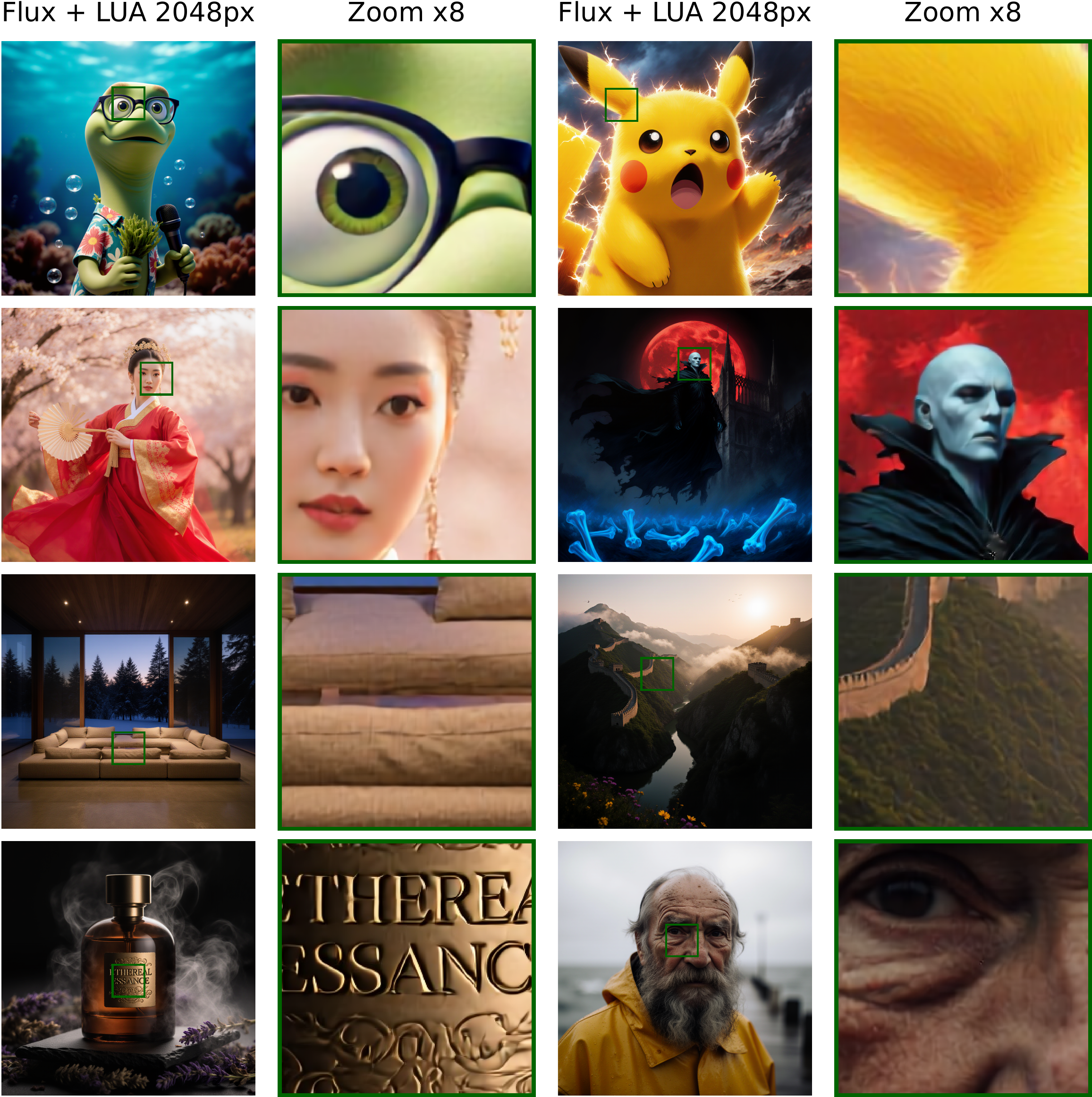}
    \caption{FLUX+LUA qualitative examples at $2048^2$ (set~2). The
    pipeline is identical to Fig.~\ref{fig:flux_lua_examples_1}:
    FLUX samples a $128{\times}128$ latent with 12 denoising steps, LUA
    performs a single $2\times$ latent upscaling step, and the FLUX
    decoder produces the final $2048{\times}2048$ image. Examples cover
    diverse content (portraits, indoor scenes, complex textures) and
    illustrate that the $C{=}16$ FLUX latents provide sufficient capacity
    for LUA to add high-frequency detail with minimal runtime overhead.}
    \label{fig:flux_lua_examples_2}
\end{figure*}

To complement the cross-model experiments in Sec.~\ref{sec:experiments},
we provide additional qualitative examples for the FLUX backbone, whose
$C{=}16$-channel latents offer a rich per-location representation on
which LUA performs particularly well. In all cases
(Figs.~\ref{fig:flux_lua_examples_1}--\ref{fig:flux_lua_examples_2}),
FLUX generates a $128{\times}128$ latent with 12 denoising steps in
$46.57$\,s on a single NVIDIA L40S GPU, LUA upsamples it to
$256{\times}256$ in $0.63$\,s, and the native FLUX decoder produces the
final $2048^2$ image in $3.40$\,s, so latent upscaling adds only about one
second of overhead beyond decoding while enhancing spatial detail.

\end{document}